\documentclass{article} 
\usepackage{iclr2025_conference, times}

\usepackage{amsmath,amsfonts,bm}

\def\eqref#1{equation~\ref{#1}}

\def\1{\bm{1}}

\DeclareMathAlphabet{\mathsfit}{\encodingdefault}{\sfdefault}{m}{sl}
\SetMathAlphabet{\mathsfit}{bold}{\encodingdefault}{\sfdefault}{bx}{n}

\newcommand{\R}{\mathbb{R}}

\usepackage{hyperref}
\usepackage{url}

\usepackage{graphicx}
\usepackage{cleveref}
\usepackage{amsmath}
\usepackage{algorithm}
\usepackage{algpseudocode}
\usepackage{array}
\usepackage{setspace}
\usepackage{booktabs}
\usepackage{multirow}
\usepackage{adjustbox}

\usepackage{authblk}

\title{MADCluster: Model-agnostic Anomaly Detection with Self-supervised Clustering Network}

\author[1]{Sangyong Lee\textsuperscript{\thanks{Corresponding author}}\,\,}  
\author[2]{Subo Hwang}
\author[3]{Dohoon Kim}

\affil[1]{AI Research Center, OKESTRO Co., Ltd.\\ \texttt{sangyong1996@gmail.com}}
\affil[2]{Seoul National University\\ \texttt{sbhwang@snu.ac.kr}}
\affil[3]{Yonsei University\\ \texttt{dkim940627@gmail.com}}

\date{}

\begin{document}
\maketitle

\begin{abstract}

This study introduces MADCluster, a model-agnostic anomaly-detection framework that leverages self-supervised clustering. MADCluster is applicable to various deep learning architectures and addresses the `hypersphere collapse' problem inherent in existing deep learning-based anomaly detection methods. The core idea is to cluster normal pattern data into a `single cluster' while simultaneously learning the cluster center and mapping data close to this center. Also, to improve expressiveness and enable effective single clustering, we propose a new `One-directed Adaptive loss'. The optimization of this loss is mathematically proven. MADCluster consists of three main components: Base Embedder capturing high-dimensional temporal dynamics, Cluster Distance Mapping, and Sequence-wise Clustering for continuous center updates. Its model-agnostic characteristics are achieved by applying various architectures to the Base Embedder. Experiments on four time series benchmark datasets demonstrate that applying MADCluster improves the overall performance of comparative models. In conclusion, the compatibility of MADCluster shows potential for enhancing model performance across various architectures.

\end{abstract}

\section{Introduction}

In modern infrastructures such as industrial equipment and data centers, numerous sensors operate continuously, generating and collecting substantial amounts of continuous measurement data. Effective detection of abnormal system patterns through real-time monitoring in these large-scale systems helps prevent significant monetary losses and potential threats \citep{djurdjanovic_watchdog_2003, leon_application_2007, yang_pipeline_2021}. However, detecting anomalies in complex time-series systems is challenging due to factors such as the diversity of abnormal patterns (irregular, unusual, inconsistent, or missing data) \citep{ruff_unifying_2021}, temporal dependencies of adjacent data, and the complexity where boundaries between normal and abnormal can be ambiguous \citep{yang_pipeline_2021}. Moreover, anomalies are generally rare, making it difficult to obtain labels and thus challenging to apply supervised or semi-supervised learning methods \citep{yang_long-distance_2021}. To tackle these challenges, a variety of time-series anomaly-detection methods have been proposed. In unlabeled environments, unsupervised learning is primarily used over supervised and semi-supervised learning. Traditional unsupervised learning-based methods include density estimation methods \citep{parzen_estimation_1962, bishop_novelty_1994, breunig_lof_nodate}, kernel-based methods \citep{scholkopf_estimating_2001, tax_support_2004}, while deep learning-based unsupervised methods include clustering-based \citep{zong_deep_2018} and deep one-class classification-based approaches \citep{ruff_deep_2018, hojjati_dasvdd_2023, shen_timeseries_2020}.

Deep one-class classification-based methods learn normal patterns of complex high-dimensional data and identify the boundaries of normal data in feature space. The main goal of these methods is to find a minimum volume region (e.g., hypersphere or hyperplane) that contains normal data, thereby detecting anomalies as data points that fall outside the learned boundary. These unsupervised anomaly detection algorithms are gaining attention due to their powerful representation learning capabilities for complex high-dimensional data and their ability to effectively model the distribution of normal data. From the standpoint of integrating multiple models to boost performance, one-class classification methods can serve as model-agnostic approaches that seamlessly adapt to diverse architectures. For example, in Log Anomaly detection tasks, they are used as an objective function to map embeddings of normal data near the normal center (\citep{guo_logbert_2021}, \citep{almodovar_logfit_2024}). However, these methods may face the ‘hypersphere collapse’ problem, a persistent issue in one-class classification where network weights converge to a trivial solution of all zeros. This leads to the problem of falling into local optima rather than global optima due to the limited expressiveness of weights in the feature space.

In this paper, we propose the \textbf{M}odel-agnostic \textbf{A}nomaly \textbf{D}etection with self-supervised \textbf{Cluster}ing network called \textbf{MADCluster}, which is applicable to existing deep learning anomaly detection models and solves the hypersphere collapse problem. The core idea of MADCluster is to cluster normal pattern data into a single cluster while simultaneously learning the cluster center and mapping data close to this center. This is motivated by the desire to achieve model-agnostic characteristics without constraints on expressiveness in the feature space. We propose an approach composed of two main modules: a distance mapping module and a clustering module. The first is a distance mapping module for mapping normal data near the center, and the second is a clustering module that learns central coordinates by single-clustering normal data. In particular, for the clustering module, we newly define an ‘One-directed Adaptive loss’ for effective single clustering and provide a proof of optimization for this One-directed  Adaptive loss. The main contributions of MADCluster are summarized as follows:

\begin{itemize}
\item \textit{Model-Agnostic Methodology}: MADCluster does not confine itself to one particular deep-learning backbone; instead, it cooperates smoothly with most mainstream architectures. Thanks to this model-agnostic stance, researchers are free from the usual trial-and-error of tailoring a detection scheme to each network, and they can expect comparable reliability whether the underlying encoder is convolutional, recurrent, or transformer-based. Such flexibility ultimately broadens the applicability of MADCluster across a wide range of empirical studies.

\item \textit{Preventing Hypersphere Collapse}: MADCluster sidesteps the hypersphere collapse problem by letting the cluster center evolve in tandem with the network weights, as the center adaptively updates through the network parameters and the representation space retains sufficient expressive power to distinguish normal data from genuine outliers. 

\item \textit{Optimization Proof for Single Clustering}: The proposed One-Directed Adaptive loss handles the center update and the decision boundary in one stroke. We supply a formal proof that optimizing the objective under mild assumptions ensures both numerical stability and theoretical soundness. As a result, practitioners can train the model without ad-hoc tricks while having confidence in its convergent behavior.

\item \textit{Performance on Public Datasets}: Even with its lean architecture, MADCluster already surpasses several well-cited baselines on four public time-series benchmarks. The margin of improvement suggests that the method extracts essential temporal cues with little overhead. Because the framework is modular, adding a stronger feature extractor or fine-tuning hyper-parameters should yield further gains without rewriting the core algorithm.

\end{itemize}

\section{Related Work}

\textit{Anomaly Detection.} Traditional anomaly detection methods follow an unsupervised paradigm, encompassing density-estimation techniques such as the Local Outlier Factor (LOF) \citep{breunig_lof_nodate}, kernel-based methods like One-Class SVM (OC-SVM) \citep{scholkopf_estimating_2001}, and Support Vector Data Description (SVDD) \citep{tax_support_2004}. These methods typically assume that the majority of the training data represents normal conditions, enabling the model to capture and learn these characteristics. Anomalies are detected when new observations do not conform well to the established model \citep{chen_one-class_2001, liu_svdd-based_2013, zhao_pattern_2013}. Recent deep-learning breakthroughs \citep{lecun_deep_2015, schmidhuber_deep_2015} have prompted researchers to embed neural networks’ rich representation-learning power into conventional classifiers. For example, DAGMM \citep{zong_deep_2018} combines Gaussian Mixture Model (GMM) with Deep Autoencoder, and DeepSVDD \citep{ruff_deep_2018} replaces the kernel-based feature space with a feature space learned by deep networks. However, DeepSVDD faces a significant issue known as hypersphere collapse, where the network weights converge to a trivial solution of all zeros \citep{ruff_deep_2018}. To mitigate this, modifications such as fixing the hypersphere center and setting the bias to zero have been implemented. While these measures help prevent hypersphere collapse, they can limit the overall performance and effectiveness of the algorithm.
In recent years, several studies have proposed solutions to the hypersphere collapse problem. DASVDD \citep{hojjati_dasvdd_2023} is structured as an autoencoder network. It involves fixing the hypersphere center $c$ to train the encoder and decoder, and then fixing the network parameters to learn the hypersphere center $c$ based on latent representations. This approach jointly trains the autoencoder and SVDD to update $c$. The Temporal Hierarchical One-class (THOC) model \citep{shen_timeseries_2020} updates the center coordinates by mapping multi-scale temporal embeddings at various resolutions near multiple hyperspheres, clustering features from all intermediate layers of the network. Both methods address the hypersphere collapse by updating the center $c$.

\textit{Clustering.} Clustering is a data mining technique that uncovers latent structure in large datasets. The primary goal of clustering is to group data points with similar characteristics, thereby identifying inherent patterns and structures within the data \citep{pavithra_survey_2017}. Traditional clustering methods include density-based clustering \citep{ester_density_1996, comaniciu_mean_2002} and distribution-based clustering \citep{bishop_pattern_2006}. These methods are effective when features are relevant and representative in finding clusters. However, they struggle to cluster high-dimensional complex data effectively as the dimensionality increases, leading to a decrease in the significance of distance measurements \citep{pavithra_survey_2017, ren_deep_2024}. To map complex data into a feature space conducive to clustering, many clustering methods focus on feature extraction or feature transformation, such as PCA \citep{wold_principal_1987}, kernel methods \citep{hearst_support_1998}, and deep neural networks \citep{liu_survey_2017}. Among these methods, deep neural networks represent a promising approach due to their excellent nonlinear mapping capabilities and flexibility. Deep Embedded Clustering (DEC) \citep{xie_unsupervised_2016} is a methodology that utilizes an autoencoder structure to learn low-dimensional representations of data and perform clustering based on these representations. Specifically, DEC defines a clustering objective function using soft cluster assignments and an auxiliary target distribution, optimizing network parameters and cluster centers while minimizing this function. Because DEC optimizes solely for clustering, it may distort local neighborhoods and compromise learned representation. Improved Deep Embedded Clustering (IDEC) \citep{guo_improved_2017} simultaneously optimizes clustering loss and reconstruction loss, enabling it to learn features while preserving the local structure of the data. Proposed method allows for consideration of both the overall cluster structure and local data relationships.

\section{Method}

In monitoring a system, we sequentially record $d$ measurements at regular intervals. In the context of time-series anomaly detection, we are given a set of time-series $\mathcal{X} = \{x_1, x_2, \dots, x_T\}$, where each point $x_t \in \mathbb{R}^d$ indicates the observation at time $t$. The goal is to detect anomalies in periodic observations to identify any deviations from normal behavior. Detecting anomalies in time-series systems presents challenges such as temporal dependencies and pattern diversity, which is why we focus on time-series anomaly detection in an unsupervised learning setting.

We have developed the model-agnostic anomaly detection with self-supervised clustering (MADCluster) network for unsupervised time-series anomaly detection, addressing the aforementioned hypersphere collapse problem while maintaining model-agnostic characteristics. MADCluster leverages the self-learning technique to update the center of the normal cluster, mapping data closer to the updated centroid and minimizing the hypersphere in the feature space. Proposed method, using dynamic centers instead of fixed ones, enables more diverse and richer representations in the feature space, thereby enhancing anomaly detection performance. Therefore, due to its model-agnostic design, MADCluster can be applied to various deep learning architectures to improve performance, and as a lightweight model with fewer parameters and faster computational speed, it poses minimal burden in terms of time cost.

\begin{figure*}[t!]
\begin{center}
\includegraphics[width=\textwidth]{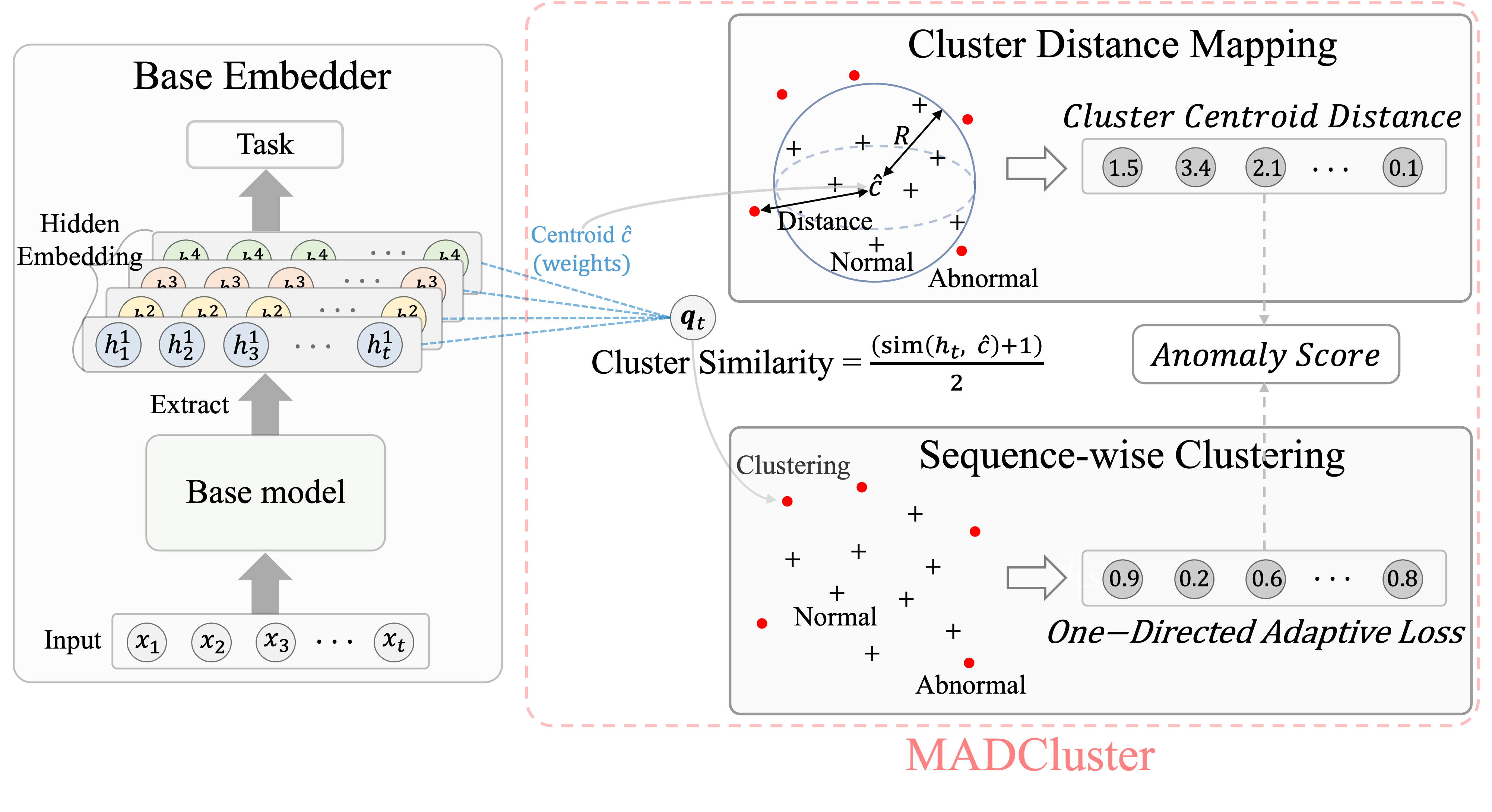} 
\vskip -0.02in
\caption{The proposed Model-agnostic Anomaly Detection with self-supervised Clustering (MADCluster) network architecture. Base Embedder captures high-dimensional temporal dynamics. Output of Base Embedder, denoted as $h_{t}$, is fed into Cluster Distance Mapping and Sequence-wise Clustering modules.}
\label{fig:MADCluster_network}
\vskip -0.18in
\end{center}
\vskip -0.15in
\end{figure*}

\subsection{Overall Architecture}

Figure~\ref{fig:MADCluster_network} illustrates the overall architecture of MADCluster, which consists of three main components: Base Embedder module, Sequence-wise Cluster module, and Cluster Distance Mapping module. On the left side, Base Embedder (\cref{Base Embedder}) initially processes the input to extract high-dimensional temporal dynamics. Extracted features are then fed into two modules on the right: Cluster Distance Mapping (\cref{Cluster Distance Mapping}) and Sequence-wise Cluster (\cref{Sequence-wise Cluster}). Cluster Distance Mapping module projects data from data space into feature space, concentrating it near the center coordinates. Sequence-wise Cluster module calculates cluster similarity for each instance and computes a One-directed Adaptive loss to update the center coordinates. Outputs of these two modules are combined through element-wise summation, which can be utilized either as an anomaly score itself or added to the anomaly score of the base model.

\subsubsection{Base Embedder}
\label{Base Embedder}

To effectively detect anomalies in time-series data, it is crucial to extract the temporal characteristics of the data well. In the Base Embedder, we use the Dilated Recurrent Neural Network (D-RNN) \citep{chang_dilated_2017} as the base model, which is designed to efficiently extract multi-scale temporal features from the time series data. D-RNN employs skip connections and dilated convolutions, allowing it to capture long-term dependencies and diverse temporal patterns across different time scales. The base model is not limited; it can utilize other anomaly detection models as well, all of which aim to extract complex hidden temporal dynamics within the data. When we consider a scenario where each process handles an input time series of length $T$, denoted as $\mathcal{X} \in \mathbb{R}^{d \times T}$, the extracted dynamics are formalized as follows:

\begin{equation} \label{eqn:base_embedder} 
    h_t = \mathcal{F}_{\text{base\_model}}(x_{t}),
\end{equation}

The output of the base model at time $t$, denoted as $h_t \in \mathbb{R}^{f \times 1}$, where $f$ represents dimensionality of the hidden feature space, reflects the learned features and extracted temporal dynamics. This flexible approach allows for the use of various models that can effectively capture the underlying temporal patterns in the data.

\subsubsection{Cluster Distance Mapping}
\label{Cluster Distance Mapping}

The MADCluster measures the deviation of the high-dimensional temporal dynamics $h_t$ from the cluster center $\hat{c}$. Unlike DeepSVDD, where the center is a pre-determined fixed point, MADCluster considers $\hat{c}$ as a learnable parameter. The objective for Cluster Distance Mapping is expressed as follows:

\begin{equation} \label{eqn:MADCLuster}
    \mathcal{L}_{\text{distance}} = R^2 + \frac{1}{\rho} \sum_{t=1}^{T} \max\left\{0, \left\|\mathrm{NN}(x_t; \mathcal{W}) - \hat{c}\right\|^2 - R^2\right\}  + \lambda\Omega(\mathcal{W}).
\end{equation}

In this case, $\mathrm{NN}(x_t; \mathcal{W})=h_t$, where $\mathrm{NN}(\cdot; \mathcal{W})$ represents a Base Embedder with parameters $\mathcal{W}$. $\Omega(\mathcal{W})$ is a regularizer (such as the $l_2$-regularizer) and \(\rho \in (0, 1]\) is a hyperparameter that balances the penalties against the sphere volume. $R$ is the radius and $\lambda$ is the learning rate. $R$ is determined based on the neural network output and the given hyperparameter $\rho$, rather than being a parameter. Instead, $R$ is computed using a specific quantile of the neural network outputs and the data loss values.

The goal is to minimize the distance loss function $\mathcal{L}_{\text{distance}}$ with respect to the neural network weights $\mathcal{W}$ and the cluster center parameters $\hat{c}$. If $\mathcal{L}_{\text{distance}}$ is updated without updating the center coordinates $\hat{c}$ through Sequence-wise Clustering, it may lead to hypersphere collapse. To mitigate this issue, MADCluster utilizes Sequence-wise Clustering to update $\hat{c}$, ensuring a continuously evolving centroid that accurately reflects the `normal' data distribution. The cluster center can be viewed as the parameters that the Sequence-wise Clustering network needs to learn. The learning process is designed to ensure that each temporal feature embedding is closely mapped to the cluster center.

\subsubsection{Sequence-wise Clustering}
\label{Sequence-wise Cluster}

In our Sequence-wise Clustering approach for anomaly detection in time-series data, we primarily focus on a single cluster representing `normal' data. Data points are classified as normal if they exhibit a high similarity of belonging to this cluster, and abnormal otherwise. While our method shares similarities with DEC \citep{xie_unsupervised_2016} in its use of self-learning for soft assignment, it diverges significantly in its approach to single clustering. Unlike conventional DEC, we discard the student's $t$-distribution, instead employing cosine similarity and a one-directed threshold to generate labels for single clustering. When the number of clusters is $k$, the clusters are denoted as $\{\hat{c}_j \in \R^f\}_{j=1}^k$. For scenarios with a single cluster center ($k=1$), we avoid using the student's $t$-distribution. In a single-cluster scenario typical of anomaly detection tasks, the student's $t$-distribution would yield a constant similarity value of 1, resulting in ineffective learning of the cluster centroid. By modifying the similarity function for soft assignment, our Sequence-wise Clustering method enables a more focused approach on the single cluster representing normal data.

Sequence-wise Clustering conducts soft assignment and auxiliary target assignment. Soft assignment calculates a cluster auxiliary distribution for each temporal feature embedding. Then, auxiliary target assignment assigns cluster labels based on a learnable one-directed threshold parameter. Sequence-wise Clustering actively performs the learning process by comparing target labels with the auxiliary distribution, in order to train closely with the normal cluster.

\textbf{Step 1 (Soft Assignment):} We used cosine similarity as the metric to compare high-dimensional temporal dynamics $h_t$ from Base Embedder with the centroid vector $\hat{c} \in \mathbb{R}^{f \times 1}$, where $\hat{c}$ is a learnable parameter. This decision enables effective centroid learning and enables our model to differentiate between normal and abnormal data in a simplified single cluster approach. The cosine similarity between high-dimensional temporal dynamics $h_t$ at time $t$ and the centroid vector $\hat{c}$ is computed as:

\begin{equation} \label{eqn:ano_q}
q_t = \frac{(h_t)^\top \cdot \hat{c}}{\|h_t\| \|\hat{c}\|},
\end{equation}

$q \in \mathbb{R}^{T \times 1}$ indicates the soft assignment similarity, and $q_t$ is subsequently normalized to a range of $0 \leq q_t \leq 1$, through the transformation $q_t = \frac{q_t + 1}{2}.$ 

\textbf{Step 2 (Auxiliary Target Assignment):} The soft assignment similarity $q_t$ is normalized and then classified into binary categories based on a one-directed threshold $\nu$ to obtain the auxiliary target. The auxiliary target is calculated as follows:

\begin{equation} \label{eqn:Auxiliary_Target_Assignment}
    \quad\quad\quad\quad\quad\quad\quad\quad p_t = 
    \begin{cases}
    1 & \text{if } q_t \geq \nu, \\
    0 & \text{otherwise,}
    \end{cases} 
    \quad\quad \text{s.t.} \quad 0 < \nu < 1
\end{equation}

$p \in \mathbb{R}^{T \times 1}$ plays the role of actual labels, and cluster center $\hat{c}$ and one-directed threshold $\nu$ are trained according to the difference between the similarity of belonging to the normal cluster, represented by $q_t$, and the auxiliary distribution $p_t$. 

\textbf{One-directed Adaptive loss function:}
We introduce a novel loss function called the One-directed Adaptive loss function. Through this proposed loss function, the one-directed threshold $\nu$ is trained to increase in value as learning progresses. The One-directed Adaptive loss function is defined as:

\vspace{1mm} 
\begin{equation}
    \mathcal{L}_{\text{cluster}}=-\sum_{t=1}^{T} p_t\log\left[\frac{1-\nu^{1-\nu}}{1-\nu}(q_t-1)+1\right]+(1-p_t)\log\left[q_t^{1-\nu}\right].
\end{equation}
\vspace{1mm} 

The One-directed Adaptive loss function has the following characteristics: First, when the value of $q_t$ is fixed, the value of $\nu$ must increase to reduce the total loss, meaning the threshold increases as it is learned. Second, the distribution of $q_t$ should approach 1, not 0, during the learning process. Calculating the derivatives $\frac{\partial \mathcal{L}_{\text{cluster}}}{\partial q_t}$ and $\frac{\partial \mathcal{L}_{\text{cluster}}}{\partial \nu}$ shows that the loss $\mathcal{L}_{\text{cluster}}$ decreases as $q_t$ and $\nu$ increase, and a detailed explanation of this is provided in~\cref{appendix:Proof of the One-directed Adaptive loss function}.

\textbf{Objective Function}: In MADCluster, the total objective function is a sum of the losses from Cluster Distance Mapping and Sequence-wise Clustering, and it is defined as follows:

\vspace{1mm} 
\begin{equation} \label{eqn:loss_FINAL}
    \mathcal{L}_{\text{total}} = \mathcal{L}_{\text{distance}} + \mathcal{L}_{\text{cluster}}.
\end{equation}

The entire procedure is detailed in Algorithm~\ref{alg:madcluster}.

\begin{algorithm}
\caption{\textbf{M}odel-agnostic \textbf{A}nomaly \textbf{D}etection with self-supervised \textbf{Cluster}ing network}
\label{alg:madcluster}
\begin{spacing}{1.2} 
\begin{algorithmic}[1]
\Require time-series $\mathcal{X} = \{x_1, x_2, \ldots, x_T\}$
\Repeat
\For{each time step $t$ in $\mathcal{X}$}
\State Process $x_t$ using Base Embedder to get $h_t$
\State Compute cosine similarity $q_t$ between $h_t$ and $\hat{c}$
\State Normalize $q_t$ to range [0, 1]
\State Assign auxiliary target $p_t$ by thresholding $q_t$ with $\nu$
\EndFor
\State Compute $\mathcal{L}_{\text{distance}}$
\State Compute $\mathcal{L}_{\text{cluster}}$
\State Set $\mathcal{L}_{\text{total}} = \mathcal{L}_{\text{distance}} + \mathcal{L}_{\text{cluster}}$
\State Update $\mathcal{W}$, $\hat{c}$, and $\nu$ based on $\mathcal{L}_{\text{total}}$ using backpropagation
\Until{convergence}
\end{algorithmic}
\end{spacing}
\end{algorithm}

\setlength{\abovedisplayskip}{3pt}

\textbf{Anomaly Score}: For a given time-series $\mathcal{X}$, consider an unseen observation at time $t$, denoted as $x_t$. The anomaly score is defined as:

\vspace{1mm} 
\begin{equation} \label{eqn:anomaly_score}
\begin{aligned}
    \mathrm{Anomaly\,\,Score}(x_t) = & -\left\{p_t\log\left[\frac{1-\nu^{1-\nu}}{1-\nu}(q_t-1)+1\right]+(1-p_t)\log\left[q_t^{1-\nu}\right]\right\}  \\
    & + \left\lVert h_{t} - c^{*}\right\rVert^2 - R^2.
\end{aligned}
\end{equation}
\vspace{1mm} 

In this case, $c^{*}$ represents the cluster center of the trained model, and $\mathrm{Anomaly\,\,Score}(x_t) \in \mathbb{R}^{T \times 1}$ serves as the point-wise anomaly score for $\mathcal{X}$. The anomaly threshold is determined using the percentile method based on the distribution of anomaly scores. Specifically, we set the threshold as the $(100-\alpha)$-th percentile of the anomaly scores, where $\alpha$ is the expected anomaly ratio. An observation $x_t$ is labeled as abnormal if $\mathrm{Anomaly\,\,Score}(x_t)$ exceeds anomaly threshold, and normal otherwise.

\begin{figure*}[t]
\vskip -0.1in
\begin{center}
\includegraphics[width=\textwidth]{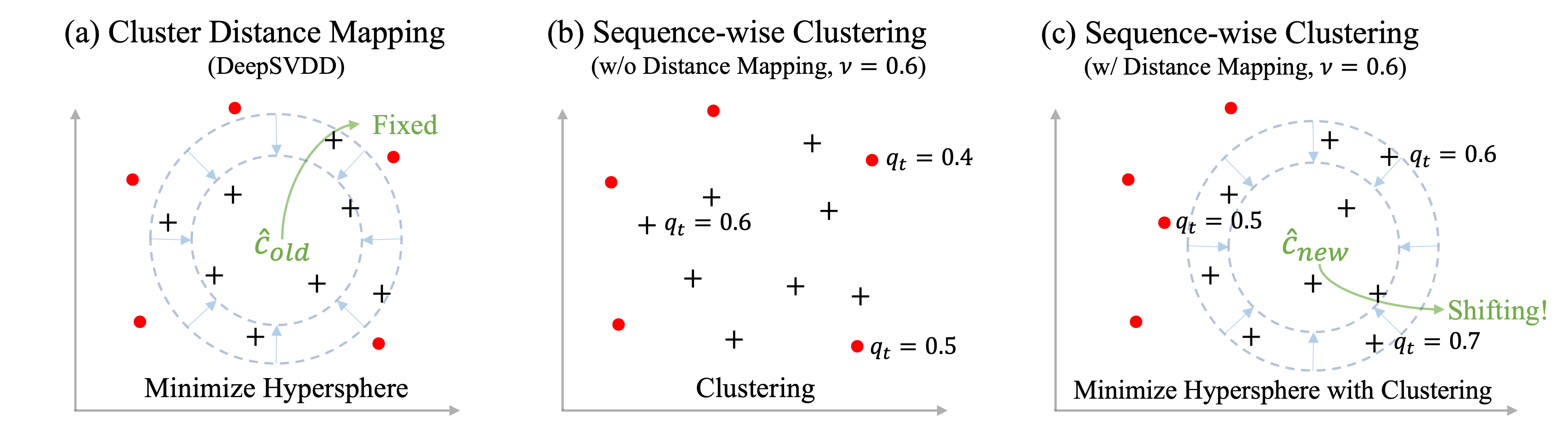} 
\vskip -0.04in
\caption{Comparison of anomaly detection approaches: (a) Cluster Distance Mapping, (b) Sequence-wise Clustering without Distance Mapping, and (c) Proposed approach combining Cluster Distance Mapping and Sequence-wise Clustering.}
\label{fig:Difference_between_method8}
\end{center}
\vskip -0.15in
\end{figure*}

Finally, to provide an intuitive understanding of the mechanism behind our proposed method, Figure~\ref{fig:Difference_between_method8} illustrates the key differences between our approach and existing techniques. This visual comparison demonstrates how our method integrates the strengths of both Cluster Distance Mapping and Sequence-wise Clustering, addressing the limitations of each approach. Red dots represent potential anomalies, black plus-sign are normal data points, and the blue circle indicates the learned hypersphere.

\begin{enumerate}
    \item \textbf{Cluster Distance Mapping (DeepSVDD):} This method employs a fixed center coordinate $\hat{c}_{\text{old}}$ and minimizes the hypersphere radius $R$ to map data points close to that center. Although this shrinking process encloses points around the fixed center, it can trap the data in a potentially suboptimal region of the feature space.
    
    \item \textbf{Sequence-wise Clustering (without Distance Mapping):} This method computes the similarity $q_t$ between the Base Embedder output $h_{t}$ and the center coordinate $\hat{c}$, then performs labeling based on a threshold $\nu$. Data points with similarity $q_t$ below the threshold are classified as anomalies. As shown, anomalies are scattered sporadically, indicating that this approach fails to capture local information effectively, potentially leading to inconsistent labeling of similar data points.
    
    \item \textbf{Combined Cluster Distance Mapping and Sequence-wise Clustering:} By integrating both approaches, our method offers key advantages. The center $\hat{c}_{\text{old}}$ is updated to $\hat{c}_{\text{new}}$ with richer representational power, and the hypersphere is minimized around this new center. Simultaneously, the algorithm leverages local information so that similar points receive consistent labels. Unlike the sporadic anomaly placements in (b), our integrated approach in (c) reflects local structure and produces more coherent anomaly predictions within similar regions of the data space.
\end{enumerate}

\section{Experiments}

\subsection{Datasets}
Description of the five experiment datasets: (1) PSM (Pooled Server Metrics, \citep{abdulaal_practical_2021}) is collected internally from multiple application server nodes at eBay with 26 dimensions. (2) Both MSL (Mars Science Laboratory rover) and SMAP (Soil Moisture Active Passive satellite) are public datasets from NASA \citep{hundman_detecting_2018} with 55 and 25 dimensions respectively, which contain telemetry anomaly data derived from the Incident Surprise Anomaly (ISA) reports of spacecraft monitoring systems. (3) SMD (Server Machine Dataset, \citep{su_robust_2019}) is a 5-week-long dataset collected from a large Internet company, consisting of 38 dimensions. The statistical details of the five benchmark datasets are summarized in Table~\ref{table:benchmark_datasets_appendix} in~\cref{appendix:Dataset}.

\subsection{Implementation Details}
Following the established protocols as outlined in previous studies \citep{shen_timeseries_2020, xu_anomaly_2021}, with a fixed window size of 100 for all datasets. Anomalies are identified among time points when their anomaly score, as defined in Equation \eqref{eqn:anomaly_score}, exceeds a specific threshold denoted as $\delta$. Importantly, we do not apply Point Adjustment (PA) in our evaluation process. Although PA is a common practice, recent studies \citep{kim2022rigorous, wang2023drift} have pointed out that PA can severely bias evaluation results. Specifically, PA utilizes ground-truth labels from the test datasets to adjust model outputs, which can lead to an overestimation of model performance. This makes fair comparison between methods impossible and undermines the validity of conclusions drawn using PA-adjusted scores. Therefore, we strictly rely on alternative evaluation metrics without PA, such as the original F1 score, threshold- and parameter-independent metrics based on VUS and Range-AUC frameworks \citep{paparrizos2022volume}, and affiliation-based Recall and Precision \citep{huet2022local}, ensuring a more rigorous and fair assessment of anomaly detection performance. 

During the experiments conducted for MADCluster, we addressed over-confidence in the output $p_t$ resulting from Sequence-wise Clustering by applying label-smoothing. The smoothing process modifies the original label $p_t$ by applying a factor $\tau$ which serves to soften the label. The softened label $p_t$ is computed using the formula $p_t = p_t \times (1-\tau) + (1 - p_t) \times \tau$. In this context, $\tau$ is the smoothing factor that is constrained by the condition $0 \leq \tau \leq 0.5$, facilitating the transition of $p_t$ from a hard to a soft label. We extensively compare our model with 11 baselines, including the reconstruction based models: USAD \citep{audibert_usad_2020}, Anomaly Transformer \citep{xu_anomaly_2021}, DCdetector \citep{yang2023dcdetector}; the clustering based methods: DeepSVDD \citep{ruff_deep_2018}, THOC \citep{shen_timeseries_2020}.

\subsection{Quantitative Results}

Table~\ref{table1:results_main_madcluster} presents the evaluation results before and after applying \textit{MADCluster} to seven baseline models across four real-world datasets: MSL, SMAP, SMD, and PSM. The results demonstrate that MADCluster consistently enhances detection performance across most evaluation metrics, with noticeable variation depending on the dataset. To ensure a comprehensive and fair evaluation of anomaly detection, we adopt a diverse set of metrics. While the original F1-score is a widely used point-wise metric, it often fails to capture the temporal continuity of anomalies. To address this limitation, we include affiliation-based metrics (\textbf{Aff-P} and \textbf{Aff-R}), which assess the spatial proximity between predicted and ground-truth anomaly regions. We also report Range-AUC-based metrics (\textbf{R\_A\_R} and \textbf{R\_A\_P}), which measure region-level alignment, and Volume-under-the-Surface metrics (\textbf{V\_ROC} and \textbf{V\_PR}), which are parameter- and threshold-free metrics designed for region-based anomaly evaluation.

On the MSL dataset, which includes structured space system telemetry anomalies, MADCluster led to a $+5.4\%p$ increase in F1, along with improvements in Aff-P ($+4.6\%p$) and V\_ROC ($+4.0\%p$), reflecting enhanced localization and discrimination capabilities. In the SMAP dataset, the F1-score improved by $+6.0\%p$, with consistent gains across all metrics. In particular, R\_A\_R and R\_A\_P increased by $+3.6\%p$, demonstrating MADCluster's strength in region-level alignment. The SMD dataset, known for its complex and noisy industrial sensor signals, saw the largest improvements, including $+6.7\%p$ in F1 and $+7.0\%p$ in Aff-P, indicating strong adaptability to high-dimensional anomaly patterns. Although the gain in Aff-R was moderate ($+1.0\%p$), increases in V\_PR and V\_ROC indicate more confident and stable detection. In contrast, the PSM dataset showed moderate improvements in F1 and other metrics (e.g., $+2.1\%p$ in V\_PR), and Aff-R increased by $+5.14\%p$, suggesting that MADCluster better capture the full extent of anomaly regions, potentially improving overall detection coverage. Overall, MADCluster proves to be an effective enhancement module that improves both precision and region-based metrics across diverse datasets, showing strong potential as a plug-in component for time series anomaly detection systems.

\begin{table}[htbp]
\label{table1:results_main_madcluster}
\centering
\scriptsize
\begin{tabular*}{\textwidth}{@{\extracolsep{\fill}}cc|ccccccc}
\toprule
\textbf{Dataset} & \textbf{Model} & \textbf{F1} & \textbf{Aff-P} & \textbf{Aff-R} & \textbf{R\_A\_R} & \textbf{R\_A\_P} & \textbf{V\_ROC} & \textbf{V\_PR} \\
\specialrule{1pt}{1pt}{1pt}
\multirow{14}{*}{MSL}
& DeepSVDD & 0.37 & 0.63 & \textbf{0.99} & 0.63 & 0.57 & 0.63 & 0.28 \\
& \textbf{+ MADCluster} & \textbf{0.52} & \textbf{0.74} & \textbf{0.99} & \textbf{0.71} & \textbf{0.65} & \textbf{0.72} & \textbf{0.42} \\ \cmidrule(lr){2-9}
& USAD & \textbf{0.53} & 0.71 & \textbf{0.97} & 0.69 & 0.66 & 0.71 & \textbf{0.43} \\
& \textbf{+ MADCluster} & \textbf{0.53} & \textbf{0.72} & \textbf{0.97} & \textbf{0.71} & \textbf{0.67} & \textbf{0.72} & \textbf{0.43} \\ \cmidrule(lr){2-9}
& BeatGAN & 0.49 & \textbf{0.71} & \textbf{0.99} & 0.68 & \textbf{0.63} & 0.70 & 0.39 \\
& \textbf{+ MADCluster} & \textbf{0.50} & \textbf{0.71} & \textbf{0.99} & \textbf{0.69} & \textbf{0.63} & \textbf{0.71} & \textbf{0.40} \\ \cmidrule(lr){2-9}
& OmniAnomaly & 0.42 & \textbf{0.64} & \textbf{1.00} & 0.63 & \textbf{0.60} & 0.64 & 0.31 \\
& \textbf{+ MADCluster} & \textbf{0.45} & \textbf{0.64} & \textbf{1.00} & \textbf{0.66} & \textbf{0.60} & \textbf{0.67} & \textbf{0.34} \\ \cmidrule(lr){2-9}
& THOC & 0.50 & 0.75 & 0.96 & \textbf{0.70} & 0.64 & 0.71 & 0.41 \\
& \textbf{+ MADCluster} & \textbf{0.55} & \textbf{0.81} & \textbf{0.99} & \textbf{0.70} & \textbf{0.68} & \textbf{0.72} & \textbf{0.46} \\ \cmidrule(lr){2-9}
& AnomalyTransformer & 0.50 & 0.70 & \textbf{0.99} & \textbf{0.69} & \textbf{0.64} & 0.70 & \textbf{0.40} \\
& \textbf{+ MADCluster} & \textbf{0.51} & \textbf{0.71} & \textbf{0.99} & \textbf{0.69} & \textbf{0.64} & \textbf{0.71} & \textbf{0.40} \\ \cmidrule(lr){2-9}
& DCdetector & 0.28 & 0.51 & 0.86 & 0.54 & 0.49 & 0.52 & 0.19 \\
& \textbf{+ MADCluster} & \textbf{0.41} & \textbf{0.64} & \textbf{0.99} & \textbf{0.64} & \textbf{0.63} & \textbf{0.64} & \textbf{0.32} \\
\specialrule{1pt}{1pt}{1pt}
\multirow{14}{*}{SMAP}
& DeepSVDD & 0.40 & 0.70 & \textbf{0.99} & 0.67 & 0.57 & 0.67 & 0.33 \\
& \textbf{+ MADCluster} & \textbf{0.49} & \textbf{0.71} & 0.96 & \textbf{0.69} & \textbf{0.61} & \textbf{0.71} & \textbf{0.38} \\ \cmidrule(lr){2-9}
& USAD & 0.44 & \textbf{0.68} & 0.98 & \textbf{0.69} & \textbf{0.61} & 0.69 & 0.35 \\
& \textbf{+ MADCluster} & \textbf{0.45} & 0.67 & \textbf{1.00} & \textbf{0.69} & \textbf{0.61} & \textbf{0.71} & \textbf{0.36} \\ \cmidrule(lr){2-9}
& BeatGAN & 0.47 & \textbf{0.70} & \textbf{1.00} & 0.69 & \textbf{0.62} & 0.71 & \textbf{0.38} \\
& \textbf{+ MADCluster} & \textbf{0.48} & \textbf{0.70} & \textbf{1.00} & \textbf{0.70} & \textbf{0.62} & \textbf{0.72} & \textbf{0.38} \\ \cmidrule(lr){2-9}
& OmniAnomaly & 0.37 & \textbf{0.64} & \textbf{1.00} & 0.64 & 0.56 & 0.65 & 0.30 \\
& \textbf{+ MADCluster} & \textbf{0.43} & \textbf{0.64} & \textbf{1.00} & \textbf{0.69} & \textbf{0.58} & \textbf{0.71} & \textbf{0.34} \\ \cmidrule(lr){2-9}
& THOC & 0.42 & \textbf{0.66} & 0.99 & 0.67 & \textbf{0.59} & 0.68 & \textbf{0.35} \\
& \textbf{+ MADCluster} & \textbf{0.44} & \textbf{0.66} & \textbf{1.00} & \textbf{0.68} & 0.57 & \textbf{0.70} & 0.33 \\ \cmidrule(lr){2-9}
& AnomalyTransformer & 0.44 & 0.66 & \textbf{1.00} & \textbf{0.69} & \textbf{0.60} & 0.70 & 0.35 \\
& \textbf{+ MADCluster} & \textbf{0.45} & \textbf{0.67} & \textbf{1.00} & \textbf{0.69} & \textbf{0.60} & \textbf{0.71} & \textbf{0.36} \\ \cmidrule(lr){2-9}
& DCdetector & 0.21 & 0.48 & 0.81 & 0.51 & 0.39 & 0.50 & 0.13 \\
& \textbf{+ MADCluster} & \textbf{0.43} & \textbf{0.67} & \textbf{0.98} & \textbf{0.67} & \textbf{0.60} & \textbf{0.69} & \textbf{0.33} \\ 
\specialrule{1pt}{1pt}{1pt}
\multirow{14}{*}{SMD}
& DeepSVDD & 0.19 & 0.56 & 0.61 & 0.53 & 0.46 & 0.53 & 0.12 \\
& \textbf{+ MADCluster} & \textbf{0.48} & \textbf{0.83} & \textbf{0.90} & \textbf{0.63} & \textbf{0.56} & \textbf{0.64} & \textbf{0.29} \\ \cmidrule(lr){2-9}
& USAD & \textbf{0.46} & 0.80 & \textbf{0.63} & \textbf{0.63} & 0.55 & 0.62 & \textbf{0.26} \\
& \textbf{+ MADCluster} & 0.45 & \textbf{0.82} & 0.61 & \textbf{0.63} & \textbf{0.57} & \textbf{0.63} & \textbf{0.26} \\ \cmidrule(lr){2-9}
& BeatGAN & 0.50 & 0.84 & 0.62 & 0.64 & \textbf{0.56} & 0.65 & \textbf{0.31} \\
& \textbf{+ MADCluster} & \textbf{0.52} & \textbf{0.85} & \textbf{0.64} & \textbf{0.65} & 0.55 & \textbf{0.66} & \textbf{0.31} \\ \cmidrule(lr){2-9}
& OmniAnomaly & 0.48 & 0.78 & 0.63 & 0.63 & \textbf{0.57} & 0.64 & 0.28 \\
& \textbf{+ MADCluster} & \textbf{0.49} & \textbf{0.82} & \textbf{0.68} & \textbf{0.64} & 0.56 & \textbf{0.65} & \textbf{0.29} \\ \cmidrule(lr){2-9}
& THOC & 0.32 & 0.70 & \textbf{0.72} & \textbf{0.61} & 0.46 & \textbf{0.61} & 0.18 \\
& \textbf{+ MADCluster} & \textbf{0.35} & \textbf{0.72} & 0.71 & \textbf{0.61} & \textbf{0.52} & \textbf{0.61} & \textbf{0.20} \\ \cmidrule(lr){2-9}
& AnomalyTransformer & 0.51 & \textbf{0.84} & \textbf{0.65} & \textbf{0.65} & \textbf{0.59} & \textbf{0.66} & \textbf{0.31} \\
& \textbf{+ MADCluster} & \textbf{0.52} & \textbf{0.84} & \textbf{0.65} & \textbf{0.65} & \textbf{0.59} & \textbf{0.66} & \textbf{0.31} \\ \cmidrule(lr){2-9}
& DCdetector & 0.17 & 0.51 & \textbf{0.98} & 0.53 & 0.31 & 0.52 & 0.12 \\
& \textbf{+ MADCluster} & \textbf{0.29} & \textbf{0.64} & 0.72 & \textbf{0.58} & \textbf{0.42} & \textbf{0.58} & \textbf{0.16} \\
\specialrule{1pt}{1pt}{1pt}

\multirow{14}{*}{PSM}
& DeepSVDD & 0.66 & 0.67 & 0.56 & 0.52 & 0.63 & 0.51 & 0.36 \\
& \textbf{+ MADCluster} & \textbf{0.67} & \textbf{0.68} & \textbf{1.00} & \textbf{0.58} & \textbf{0.69} & \textbf{0.58} & \textbf{0.46} \\ \cmidrule(lr){2-9}
& USAD & \textbf{0.52} & \textbf{0.56} & \textbf{0.99} & 0.57 & \textbf{0.69} & \textbf{0.56} & \textbf{0.38} \\
& \textbf{+ MADCluster} & \textbf{0.52} & \textbf{0.56} & \textbf{0.99} & \textbf{0.58} & \textbf{0.69} & \textbf{0.56} & \textbf{0.38} \\ \cmidrule(lr){2-9}
& BeatGAN & 0.66 & 0.68 & 0.70 & 0.63 & 0.67 & 0.63 & \textbf{0.49} \\
& \textbf{+ MADCluster} & \textbf{0.68} & \textbf{0.74} & \textbf{0.80} & \textbf{0.67} & \textbf{0.69} & \textbf{0.66} & \textbf{0.49} \\ \cmidrule(lr){2-9}
& OmniAnomaly & \textbf{0.53} & \textbf{0.55} & \textbf{0.98} & \textbf{0.54} & \textbf{0.67} & \textbf{0.53} & \textbf{0.38} \\
& \textbf{+ MADCluster} & \textbf{0.53} & \textbf{0.55} & \textbf{0.98} & \textbf{0.54} & \textbf{0.67} & \textbf{0.53} & \textbf{0.38} \\ \cmidrule(lr){2-9}
& THOC & \textbf{0.50} & \textbf{0.55} & \textbf{1.00} & \textbf{0.52} & \textbf{0.69} & \textbf{0.51} & \textbf{0.36} \\
& \textbf{+ MADCluster} & \textbf{0.50} & \textbf{0.55} & \textbf{1.00} & \textbf{0.52} & \textbf{0.69} & \textbf{0.51} & \textbf{0.36} \\ \cmidrule(lr){2-9}
& AnomalyTransformer & 0.53 & 0.55 & \textbf{0.98} & 0.54 & \textbf{0.67} & 0.53 & 0.38 \\
& \textbf{+ MADCluster} & \textbf{0.54} & \textbf{0.56} & 0.94 & \textbf{0.55} & \textbf{0.67} & \textbf{0.54} & \textbf{0.39} \\ \cmidrule(lr){2-9}
& DCdetector & 0.50 & 0.55 & \textbf{1.00} & 0.52 & \textbf{0.69} & 0.51 & 0.36 \\
& \textbf{+ MADCluster} & \textbf{0.53} & \textbf{0.57} & 0.86 & \textbf{0.55} & 0.63 & \textbf{0.54} & \textbf{0.40} \\ \cmidrule(lr){2-9}
\end{tabular*}
\caption{Performance metrics (original F1, Aff-P, Aff-R, R\_A\_R, R\_A\_P, V\_ROC, V\_PR) for 7 models before and after applying MADCluster on four datasets. Results are in decimal format (ranging from 0 to 1), with best results in bold.}
\end{table}

\subsection{Qualitative Results}

\begin{figure*}[t]
\begin{center}
\centerline{\includegraphics[width=\textwidth]{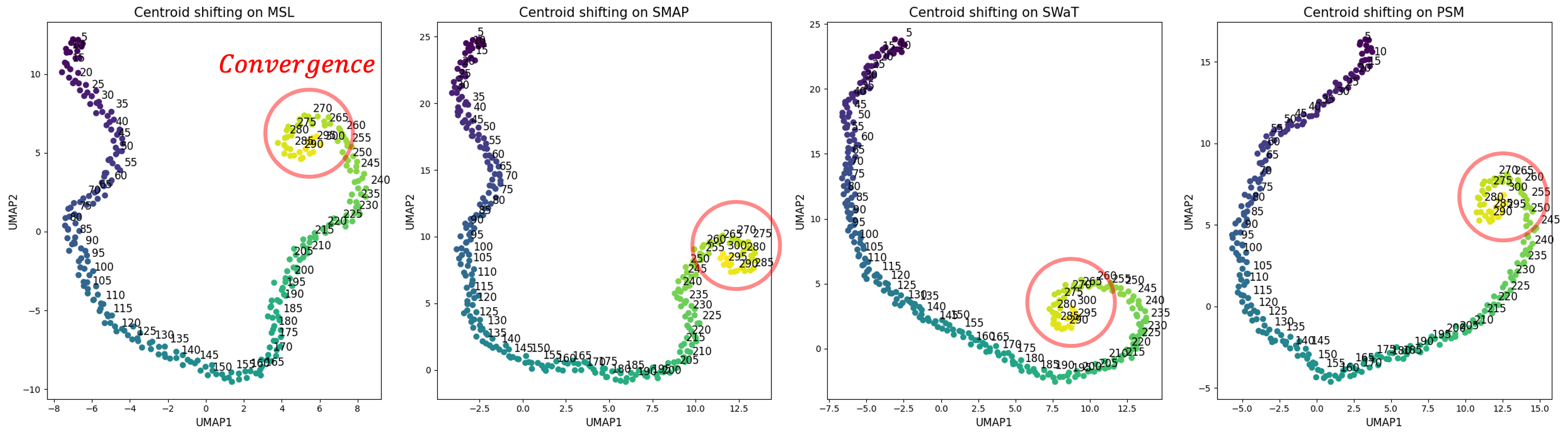}}
\caption{Visualization of centroid movement, captured every 5 epochs using UMAP.}
\label{figure:Visualization_of_centroid_movement}
\end{center}
\vskip -0.2in
\end{figure*}

\begin{figure*}[h]
\begin{center}
\centerline{\includegraphics[width=\textwidth]{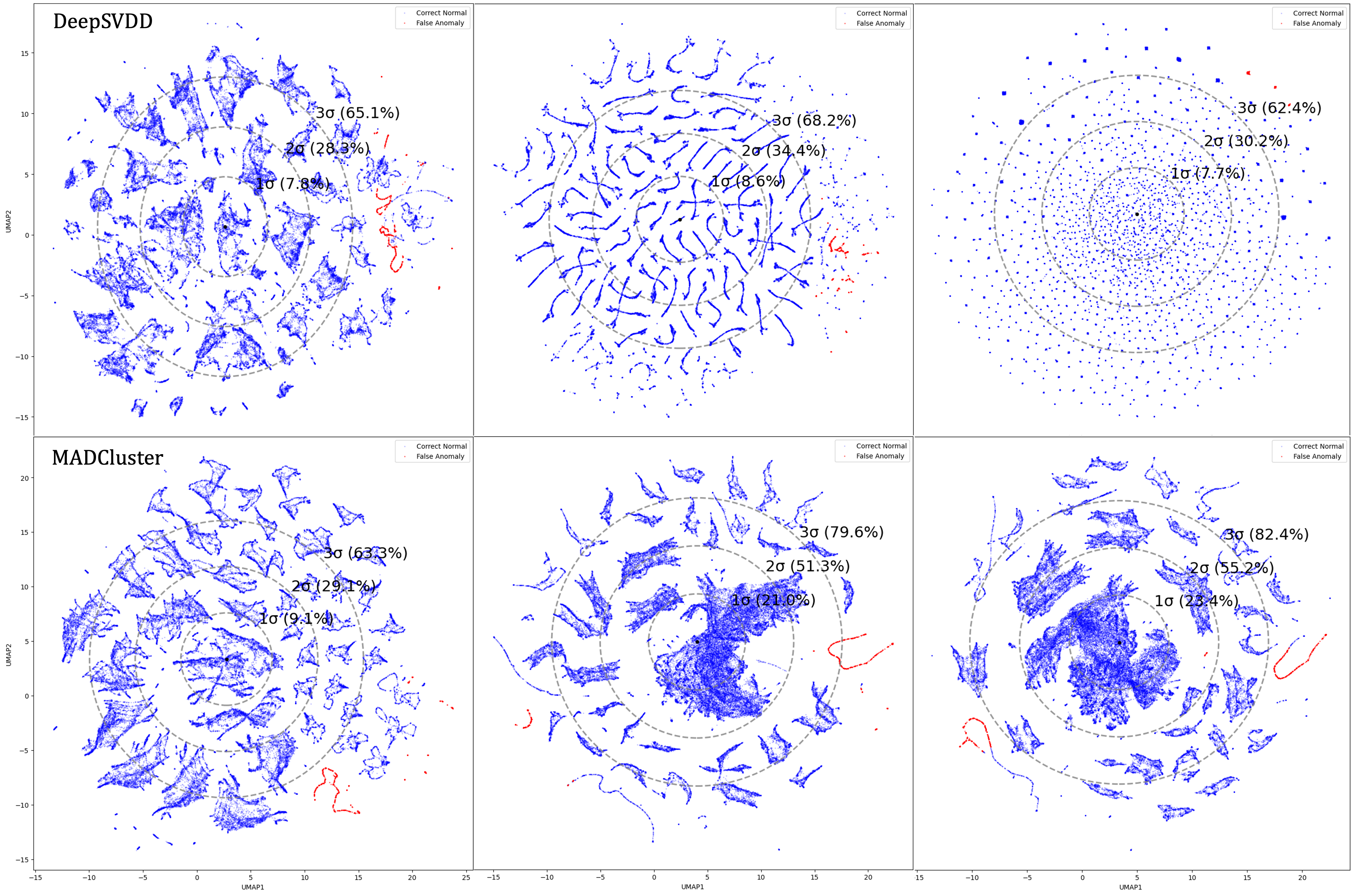}}
\caption{Hidden embedding visualization for DeepSVDD (top) and MADCluster (bottom) at epochs 1, 150, and 300. $\sigma$ represents the standard deviation from the center of hidden embeddings.}
\label{hidden_map}
\end{center}
\end{figure*}

We have addressed the limitations of previous models, particularly the issue of fixed center coordinates, through our proposed method, MADCluster. To visualize how the center coordinates move and converge, we employed UMAP \citep{mcinnes_umap_2018}, a dimensional reduction technique, to represent the high-dimensional centroid in two-dimensional space. Figure~\ref{figure:Visualization_of_centroid_movement} presents the two-dimensional mapping results across four datasets. This figure illustrates the evolution of cluster center coordinates, updated through MADCluster, visualized in two dimensions over 300 epochs. Throughout the training process, we observe that the cluster center converges towards specific points, exhibiting vibrating behavior within the converged area. This convergence, as opposed to divergence, indicates that the center coordinates are learning to represent more complex feature spaces. In Figure~\ref{hidden_map} to verify the effectiveness of the moving center coordinates during training and provide an intuitive understanding, we conducted a visual comparison between DeepSVDD and MADCluster.

\vskip +0.1in

This visualization illustrates how well the embeddings learned through each model are clustered around the center on the MSL dataset. The embeddings of each model are visualized in two dimensions after training for 1, 150, and 300 epochs. All visualized data represents normal instances only, with blue points indicating correctly classified normal data and red points showing false anomaly detections. At epoch 1, both DeepSVDD and MADCluster display a dispersed distribution of data around the center. For our proposed method, 9.1\%, 29.1\%, and 63.3\% of the data fall within 1, 2, and 3 sigma, respectively. At epoch 150, DeepSVDD exhibits a scattered distribution, while MADCluster shows data converging towards the center. MADCluster encompasses 21.0\%, 51.3\%, and 79.6\% of the data within 1, 2, and 3 sigma, demonstrating that more data points have moved closer to the center compared to the initial epoch. By epoch 300, DeepSVDD forms multi-cluster at various points away from the center, whereas MADCluster continues to draw data closer to the center. MADCluster now includes 23.4\%, 55.2\%, and 82.4\% of the data within 1, 2, and 3 sigma. MADCluster steadily draws the hidden embeddings toward a single cluster center, fulfilling its design purpose. DeepSVDD, by comparison, shows no such centripetal trend; its embeddings separate into several local groupings, leaving the feature space fragmented rather than unified. The visual evidence thus confirms that MADCluster’s adaptive center not only expands representational capacity but also averts the hypersphere-collapse risk inherent in fixed-centroid models. Because the center is updated dynamically, the method provides a more flexible and faithful representation of the normal data manifold.

\begin{figure*}[t!]
\vskip -0.05in
\begin{center}
\centerline{\includegraphics[width=\textwidth]{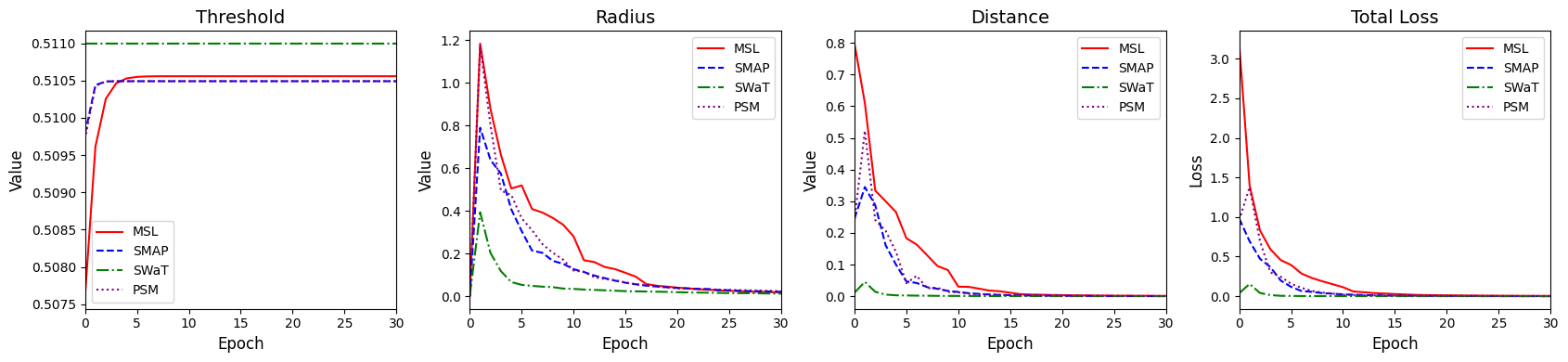}}
\vskip -0.1in
\caption{Visualization of the changes in threshold, radius, distance, and loss during training on four datasets.}
\label{loss_plot}
\end{center}
\vskip -0.3in
\end{figure*}

Figure~\ref{loss_plot} visualizes the changes in threshold, radius, distance, and loss during the training process across four datasets illustrating how each metric evolves as training progresses. The threshold, which refers to the one-directed threshold, shows a pattern of gradual increase in the early stages of training before eventually converging. After the threshold converges, radius, distance, and loss generally exhibit a decreasing trend. This pattern is consistently observed across all datasets. The proposed one-directed threshold method can serve as an indicator to assess whether the training is proceeding correctly.

\section{Conclusion and Future Work}

MADCluster is a novel, model-agnostic anomaly-detection framework that incorporates a self-supervised clustering network called \textit{MADCluster}. It can be seamlessly integrated into existing deep-learning models and mitigates the hypersphere-collapse problem. The framework is built on three modules: (i) a Base Embedder for capturing high-dimensional temporal dynamics, (ii) Cluster-Distance Mapping, which constrains embeddings near the normal cluster center, and (iii) Sequence-wise Clustering, a self-learning mechanism that continuously updates that center. Evaluated on four benchmark datasets, MADCluster produced more expressive center coordinates and yielded consistent gains in Aff-P, Aff-R, V\_ROC, and V\_PR, indicating sharper decision boundaries, more discriminative anomaly scores, and broader coverage of anomalous regions. Future studies should examine how each structural component affects recall across diverse anomaly types and explore adaptive strategies that further balance precision and recall. 

\bibliography{iclr2025_conference}
\bibliographystyle{iclr2025_conference}

\newpage  
\appendix
\section{Proof of the One-directed Adaptive loss function}
\label{appendix:Proof of the One-directed Adaptive loss function}

In this chapter, we will explain our own loss function. First, we analyze why Binary Cross Entropy (BCE) is inadequate for our situation. What we're trying to achieve serves as a clear motivation for a newly constructed loss function. Then, using the properties of a function whose exponent is a positive rational number less than 1, a new loss function is defined. In the last part of this chapter, the derivative of this loss function and the sign of the derivative are mathematically considered, to ensure that the total loss function actually decreases during the learning process. For simplicity in this Appendix, we will use $q$ and $p$ to represent $q_t$ and $p_t$ respectively, without loss of generality. This notation will be used consistently throughout the following proofs and explanations.

\subsection{Motivation for Proposing One-directed Adaptive loss}
\subsubsection{Analysis to Binary Cross Entropy}

We will first examine a brief analysis of the BCE. The loss function is constructed as follows:

\begin{equation}
    \mathcal{L}_{\text{cluster}}=-\sum p\log q+(1-p)\log (1-q)
\end{equation}

Before calculating $p$ by \eqref{eqn:Auxiliary_Target_Assignment} using one-directed threshold, assume that the threshold is fixed as 0.5 in the loss function. Then, $p$ is determined by the following rule:

\begin{equation}
    p=\begin{cases}
    0, & \;\; 0 \le q < 0.5 \\
    1, & \;\; 0.5 \le q \le 1
\end{cases}
\end{equation}

So the loss function is calculated by different functions depending on which interval the value of $q$ belongs to. In the BCE, the total interval [0, 1] for the available value of $q$ is divided by a threshold, which is 0.5, into two different intervals: [0, 0.5) and [0.5, 1]. To simplify the analysis, let's consider a function where the variable $q$ is on the $x$-axis and the value inside the logarithm is on the $y$-axis. Then we can reconstruct the original BCE into:

\begin{equation}
    \;\; \;\; y=\begin{cases}
    1-q, & \;\; 0 \le q < 0.5 \\
    q,   & \;\; 0.5 \le q \le 1
\end{cases}
\end{equation}

\begin{figure*}[h]
\vskip -0.045in
\begin{center}
\centerline{\includegraphics[width=\textwidth]{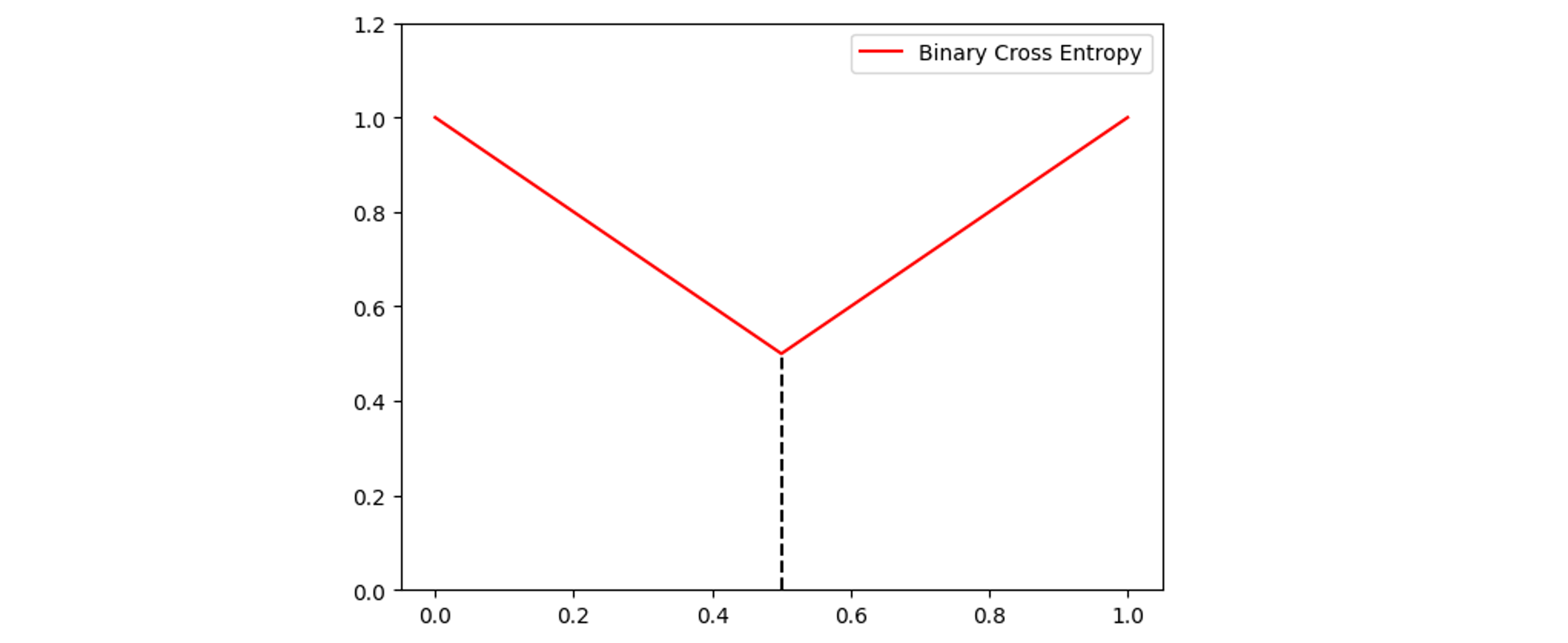}}
\caption{The black dashed line represents the position of the threshold that acts on the value $q$ to classify whether the label is 0 or 1.}
\label{bce_loss}
\end{center}
\end{figure*}

Figure~\ref{bce_loss} shows the value inside the logarithm in the BCE loss function. To reduce the total loss, the value inside the logarithm must be increased.

\vskip -0.05in

Therefore, the closer the value of $y$ is to 1, the smaller the total loss. The distribution of $q$ can therefore be classified into two different labels. One will be located in the neighborhood of 0 and the other will be located in the neighborhood of 1. However, this approach poses a problem in anomaly detection tasks using single clustering, particularly when training only on normal data. The issue arises because the BCE loss function allows normal data to be correctly classified whether it's close to 0 or 1. We typically want normal data to cluster towards one direction - either 0 or 1, not both. The learning process should encourage normal data to converge towards a single value (either 0 or 1), rather than allowing it to be distributed at both extremes.

\subsubsection{Desired Goals}

What we are aiming for requires two differences from the original loss function. The first one is that the threshold must be learned, and the threshold must increase as it is learned. And second, the distribution of $q$ should only be close to $1$, not to $0$, during the learning process. If the threshold is denoted by $\nu$, we will take a monotonic function such that the overall graph should approach $y=1$ as the value of $\nu$ increases as a value part of the logarithm of a new loss function.

\subsection{The One-directed Adaptive Loss function Modeling}

At first, the total interval $[0, 1]$ in which all possible $q$ values is divided into $[0, \nu)$ and $[\nu, 1]$. Then the value $p$ is determined as follows:

\vspace{1mm} 
\begin{equation}
    p = \begin{cases}
    0, & \;\; 0 \leq q < \nu \\
    1, & \;\; \nu \leq q \leq 1
\end{cases}
\end{equation}
\vspace{1mm} 

To avoid the situation where the loss function is not defined, assume that the possible $\nu$ is in the range $0<\nu<1$. The simplest monotonic function connecting two points $(0, 0)$ and $(1, 1)$ is of the form $y = q^n$. For $n$ which satisfies the inequality $0<n<1$, the functions $y = q^n$ are close to $y = 1$ as $n$ decreases. So consider the following function to match the increasing trend of $\nu$ with the decreasing trend of $n$:

\begin{equation}
    y = q^{1-\nu}
\end{equation}
\vspace{1mm} 

\begin{figure*}[h]
\begin{center}
\centerline{\includegraphics[width=7.5cm]{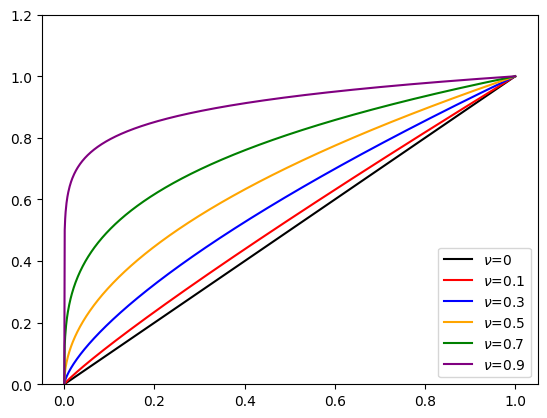}}
\caption{The graph of $y=q^{1-\nu}$ with different values of $\nu$ between $0$ and $1$. }
\label{Front_interval_graph}
\end{center}
\end{figure*}

Figure~\ref{Front_interval_graph} shows the graphs of the above function with different values of $\nu$ between $0$ and $1$. As $\nu$ increases, it can be seen that starting from $y=x$ and approaching $y=1$ rapidly. This effect is more pronounced at lower values of $q$.
\vspace{1mm} 

Qualitatively, this function is rapidly increasing to $1$ for small $q$ when $\nu$ is increasing. So we adopt the function $q^{1-\nu}$ in the interval $[0, \nu)$ as the value inside the logarithm of the loss function. Meanwhile, in the interval $[\nu, 1]$, we define the function as a linear function connecting two points $(\nu, \nu^{1-\nu})$ and $(1, 1)$, ensuring the continuity of the entire function over the interval $[0, 1]$ and reflecting the simplest form.

\vspace{1mm} 
\begin{equation}
y=\frac{1-\nu^{1-\nu}}{1-\nu}(q-\nu)+\nu^{1-\nu}=\frac{1-\nu^{1-\nu}}{1-\nu}(q-1)+1
\end{equation}
\vspace{1mm} 

In summary, we adopt the following function as the value inside the logarithm of our new loss function.

\vspace{1mm} 
\begin{equation}
y=\begin{cases}
q^{1-\nu}, & 0 \leq q < \nu \\
\frac{1-\nu^{1-\nu}}{1-\nu}(q-1)+1, & \nu \leq q \leq 1
\end{cases}
\end{equation}
\vspace{1mm} 

\begin{figure*}[h]
\begin{center}
\includegraphics[width=7.5cm]{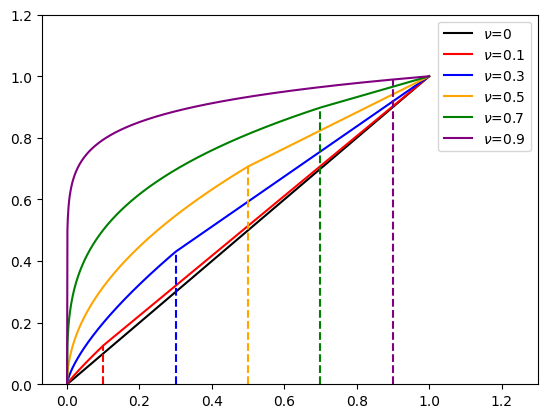}
\vskip -0.1in
\caption{The graph of our new loss function with different values of $\nu$ between $0$ and $1$.}
\label{New_loss_graph}
\end{center}
\end{figure*}

Corresponding graphs with different $\nu$ are shown in Figure~\ref{New_loss_graph}. Each colored dashed line indicates the position of the threshold at different values of $\nu$. Before the threshold, the function is concave; after it, the function is linear.

Thus, the final loss function can be expressed as follows:

\vspace{1mm} 
\begin{equation} \label{eqn:L_cluster}
    \mathcal{L}_{\text{cluster}} = -\sum p\log\left(\frac{1-\nu^{1-\nu}}{1-\nu}(q-1)+1\right) + (1-p)\log\left(q^{1-\nu}\right)
\end{equation}
\vspace{1mm} 

\subsection{Derivative of loss function}
In order to mathematically confirm that the new loss function really decreases when $q$ and $\nu$ are increasing, to simplify the derivative procedure, let us define $f_1$ and $f_2$ as:

\vspace{1mm} 
\begin{equation}
f_1\equiv\frac{1-\nu^{1-\nu}}{1-\nu}(q-\nu)+\nu^{1-\nu}=\frac{1-\nu^{1-\nu}}{1-\nu}(q-1)+1,\qquad f_2\equiv q^{1-\nu}
\end{equation}
\vspace{1mm} 

Since both $f_1$ and $f_2$ satisfy the conditions for a valid logarithm argument, $f_1$ and $f_2$ are positive in the entire interval $[0, 1]$. The derivative of total loss $\mathcal{L}_{\text{cluster}}$ with respect to $q$ and $\nu$ can be expressed as:

\begin{equation}
    \frac{\partial \mathcal{L}_{\text{cluster}}}{\partial q}=-p\frac{1}{f_1}\frac{\partial f_1}{\partial q}-(1-p)\frac{1}{f_2}\frac{\partial f_2}{\partial q},
    \quad\quad
    \frac{\partial \mathcal{L}_{\text{cluster}}}{\partial \nu}=-p\frac{1}{f_1}\frac{\partial f_1}{\partial \nu}-(1-p)\frac{1}{f_2}\frac{\partial f_2}{\partial \nu}.
\end{equation}

\vspace{1mm} 
\vspace{1mm} 
\vspace{1mm} 
\subsubsection{$\partial \mathcal{L}_{\text{cluster}} / {\partial q}$}
\vspace{1mm} 

Since both $f_1$ and $f_2$ are positive, we need to verify the signs of $\partial f_1/\partial q$ and $\partial f_2/\partial q$. Let's consider the derivative of $f_1$ with respect to $q$ first:

\vspace{1mm} 
\begin{equation}
    \frac{\partial f_1}{\partial q}=\frac{1-\nu^{1-\nu}}{1-\nu}
\end{equation}
\vspace{1mm} 

The condition $0<\nu<1$ implies $0<\nu^{1-\nu}<1$. Therefore, both the denominator and the numerator are positive, ensuring that $\partial f_1/\partial q>0$ is satisfied.
Meanwhile, the derivative of $f_2$ with respect to $q$ can be written as:

\vspace{1mm} 
\begin{equation}
    \frac{\partial f_2}{\partial q}=(1-\nu)q^{-\nu}=\frac{1-\nu}{q^\nu}
\end{equation}
\vspace{1mm} 

Similarly, because $0<\nu<1$ and $0<q<1$, both the denominator and the numerator are also positive, so $\partial f_2/\partial q>0$ is satisfied.
Thus, we can determine the sign of the derivative of our new loss function with respect to $q$:

\vspace{1mm} 
\begin{equation}
    \frac{\partial \mathcal{L}_{\text{cluster}}}{\partial q}<0
\end{equation}
\vspace{1mm} 

This means that the total loss $\mathcal{L}_{\text{cluster}}$ decreases as $q$ increases.

\vspace{1mm} 
\vspace{1mm} 
\vspace{1mm} 
\subsubsection{$\partial \mathcal{L}_{\text{cluster}} / {\partial \nu}$}
\vspace{1mm} 

This part is very similar to proving the sign of $\partial\mathcal{L}_{\text{cluster}}/\partial q$, but it requires a more technical procedure. The derivative of total loss $\mathcal{L}_{\text{cluster}}$ with respect to $\nu$ can be written as follows:

\vspace{1mm} 
\begin{equation}
    \frac{\partial \mathcal{L}_{\text{cluster}}}{\partial \nu}=-p\frac{1}{f_1}\frac{\partial f_1}{\partial \nu}-(1-p)\frac{1}{f_2}\frac{\partial f_2}{\partial \nu}
\end{equation}
\vspace{1mm} 

Since both $f_1$ and $f_2$ are positive, we need to verify the signs of $\partial f_1/\partial \nu$ and $\partial f_2/\partial \nu$. Let's consider the derivative of $f_1$ with respect to $\nu$ first:

\vspace{1mm} 
$$
    \frac{\partial f_1}{\partial \nu}=\frac{(q-1)}{(1-\nu)^2}\left[-\nu^{1-\nu}\left(\frac{1-\nu}{\nu}-\log \nu\right)(1-\nu)+(1-\nu^{1-\nu})\right]
$$

$$
    =\frac{(q-1)}{(1-\nu)^2}\left[1+\nu^{1-\nu}\left(-\frac{(1-\nu)^2}{\nu}+(1-\nu)\log \nu -1\right)\right]
$$

\begin{align}
    =\frac{(q-1)}{(1-\nu)^2\nu^\nu}\left\{\nu^\nu+\nu-\nu^2-1+\nu(1-\nu)\log \nu\right\}
\end{align}
\vspace{1mm} 

We have a condition for $c$ and $q$, which is $0<\nu<1$ and $0<q<1$. The outermost factor satisfies the following inequality:

\vspace{1mm} 
\begin{equation}
    \frac{(q-1)}{(1-\nu)^2\nu^\nu}<0
\end{equation}
\vspace{1mm} 

Let us define $g_1, g_2, g_3$ as:

\vspace{1mm} 
\begin{equation}
    \begin{cases}
        g_1=\nu^\nu+\nu \\
        g_2=\nu^2+1 \\
        g_3=\nu(1-\nu)\log \nu
    \end{cases}
\end{equation}
\vspace{1mm} 
\vspace{1mm} 
\vspace{1mm} 

To express the formula inside the braces as $g_1-g_2+g_3$, we will confirm the sign of each function for $\nu\in(0, 1)$, thereby justifying the sign of the formula inside the braces. $g_3$ satisfies $g_3<0$ because of two inequalities:

\vspace{1mm} 
\begin{equation}
    \log \nu<0, \qquad \nu(1-\nu)>0
\end{equation}
\vspace{1mm} 
\vspace{1mm} 
\vspace{1mm} 

From the limit $\lim_{\nu\rightarrow 0+}\nu^\nu=1$, we can obtain the values of $g_1$ and $g_2$ at $\nu=1$ and the left-side limit values of $g_1$ and $g_2$:

\vspace{1mm} 
\begin{equation}
    \begin{cases}
    g_1(0+)=g_2(0+)=1  \\ 
    g_1(1)=g_2(1)=2
    \end{cases}
\end{equation}
\vspace{1mm} 
\vspace{1mm} 
\vspace{1mm} 

The derivative of $g_1$ with respect to $\nu$ is:

\vspace{1mm} 
\vspace{1mm} 
\begin{equation}
    \frac{\partial g_1}{\partial \nu} =\nu^\nu(1+\log \nu)+1
\end{equation}
\vspace{1mm} 
\vspace{1mm} 

Here, the first term $\nu^\nu(1+\log \nu)$ is negative when $\nu\in(0, e^{-1})$, while it is positive due to the factor $(1+\log \nu)$ when $\nu\in(e^{-1}, 1)$. Consequently, the function $g_1-\nu$ decreases in the interval $(0, e^{-1})$ and increases in the interval $(e^{-1}, 1)$. Additionally, the first term $\nu^\nu(1+\log \nu)$ diverges to $-\infty$ as $\nu$ approaches 0 from the positive side. While the interval of increase or decrease might differ by adding the constant 1 to the first term, the overall trend of $g_1$ remains the same even when considering $g_1-\nu$. The derivative of $g_2$ with respect to $\nu$ is:

\vspace{1mm} 
\begin{equation}
    \frac{\partial g_2}{\partial \nu}=2\nu
\end{equation}
\vspace{1mm} 

This quantity is always positive if $\nu\in(0, 1)$, so the function $g_2$ increases in the interval $(0, 1)$. Therefore, in the interval $(0, 1)$, the function $g_1$ is always smaller than the function $g_2$; $g_1-g_2<0$. This means that the formula $g_1-g_2+g_3$ satisfies the following inequality where $\nu\in(0, 1)$:

\begin{equation}
    g_1-g_2+g_3<0
\end{equation}

\begin{figure*}[t]
\begin{center}
\includegraphics[width=\textwidth]{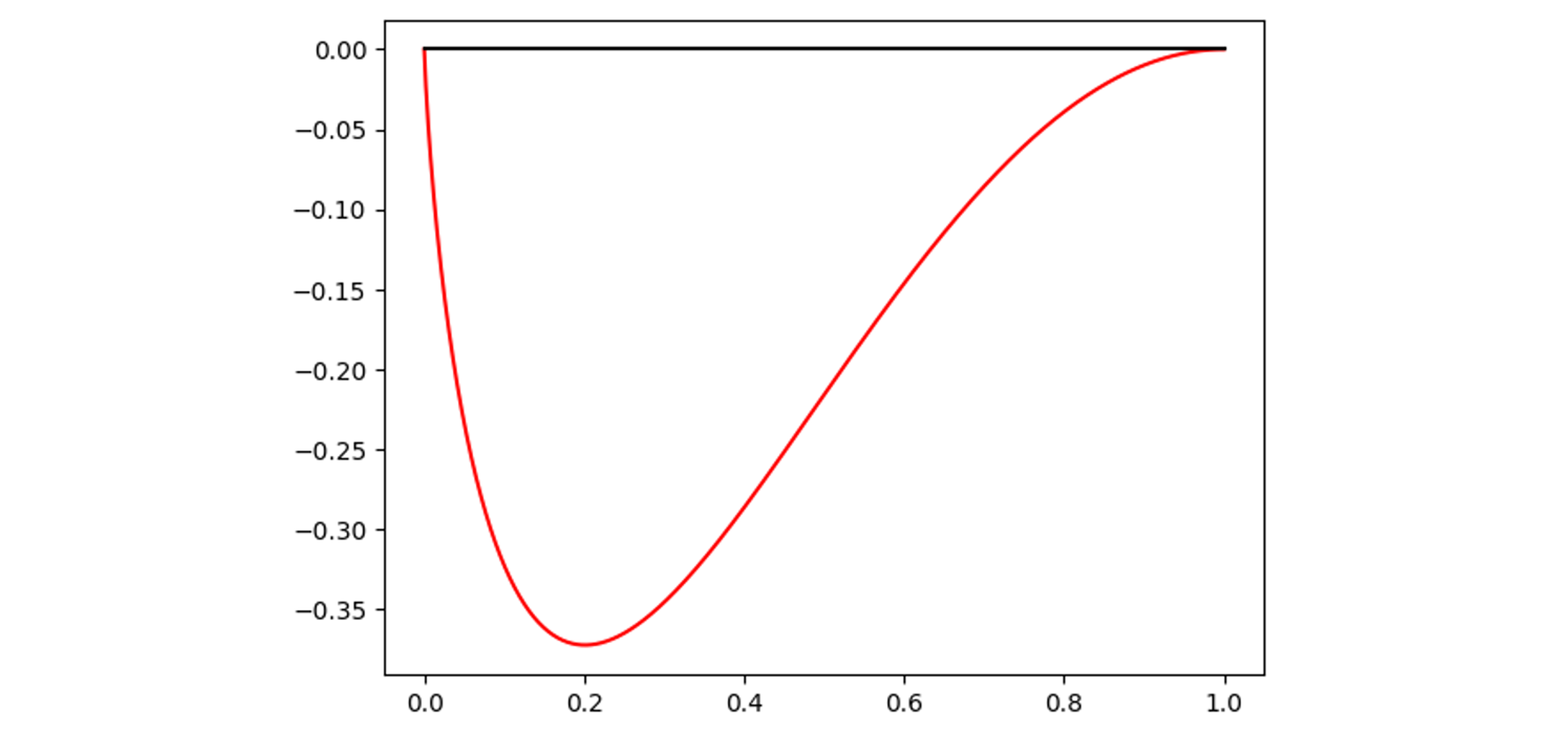}
\caption{The graph of $g_1-g_2+g_3$ in the interval $[0, 1]$. The black line represents the $x$-axis; values below this line indicate that the function is negative.}
\label{fig:g1g2g3}
\end{center}
\end{figure*}

Indeed, the graph of $g_1-g_2+g_3$ represents negative values in the interval $(0, 1)$, as shown in Figure~\ref{fig:g1g2g3}.

Therefore, the sign of the derivative of $f_1$ with respect to $\nu$ is positive, so $\partial f_1/\partial \nu>0$. On the other hand, for $\partial f_2/\partial \nu$, we have:

\begin{equation}
    \frac{\partial f_2}{\partial \nu}=-q^{1-\nu}\log q
\end{equation}

The value of $q^{1-\nu}$ is between $0$ and $1$, and $\log q<0$, so $\partial f_2/\partial \nu>0$. Thus, we can determine the sign of the derivative of our new loss function with respect to $\nu$:

\begin{equation}
    \frac{\partial \mathcal{L}_{\text{cluster}}}{\partial \nu}<0
\end{equation}

This means that the total loss $\mathcal{L}_{\text{cluster}}$ decreases as $\nu$ increases.

\section{Multi-cluster ($k>1$) for MADCluster}
\label{appendix:Multi-cluster based Results of MADCluster}

MADCluster employs cosine similarity with a One-directed Adaptive loss function, initially assuming a single cluster ($k=1$). This design overcomes the trivial solution where the soft assignment of a student's $t$-distribution always yields a value of 1 when only one cluster is present. Whereas, with several modifications, MADCluster can be extended utilizing student's $t$-distribution to support multi-cluster based clustering ($k>1$). The soft assignment \(q_{tj}\) and the target distribution \(p_{tj}\) represent the assignment of the $t$-th representation to the $j$-th cluster and is defined as:

\begin{equation} \label{eqn:q_tj}
\begin{aligned}
    q_{tj} = \frac{(1 + |h_t^{f} - \hat{c}_j|^2)^{-1}}{\sum_{j=1} (1 + |h_t^{f} - \hat{c}_j|^2)^{-1}},
    \quad\quad
    p_{tj} = \frac{q_{tj}^2 / \sum_{t=1} q_{tj}}{\sum_{j=1} (q_{tj}^2 / \sum_{t=1} q_{tj})}
\end{aligned}
\end{equation}

Sequence-wise Clustering loss \(\mathcal{L}_{\text{cluster}}\) is calculated using the Kullback-Leibler (KL) divergence instead of the One-directed Adaptive loss. It is defined as follows:

\begin{equation} \label{eqn:L_cluster_tj}
\begin{aligned}
\mathcal{L}_{\text{cluster}} = KL(P|Q) = \sum_{j=1}^{K} \sum_{t=1}^{T} p_{tj} \log \frac{p_{tj}}{q_{tj}}
\end{aligned}
\end{equation}

And for the Cluster Distance Mapping loss \(\mathcal{L}_{\text{distance}}\) we have adopted a simplified notation, omitting some details for clarity, is also defined as follows:

\begin{equation} \label{eqn:L_distance_tj}
\begin{aligned}
    \mathcal{L}_{\text{distance}} = \frac{1}{n} \sum_{j=1}^{K} \sum_{t=1}^{T} \| h_t^{f} - \hat{c}_j \|^2 + \lambda \Omega(\mathcal{W})
\end{aligned}
\end{equation}

Consequently, during training, we sum two components for each time step $t$: the KL-divergence values across all clusters for the $t$-th representation, and the distances from the $t$-th representation to each cluster center. The anomaly score is also defined as follows:

\begin{equation} \label{eqn:anomaly_score2}
\begin{aligned}
    \mathrm{Anomaly\,\,Score}(x_t) = & \sum_{j=1}^{K} p_{tj} \log \frac{p_{tj}}{q_{tj}}
    + \left\lVert h_{t}^{f} - c_{j}^{*}\right\rVert^2
\end{aligned}
\end{equation}

For the multi-cluster case, the anomaly score does not incorporate $\nu$, and therefore $\nu$ is not learned. Similar to the single-cluster case, $\mathrm{Anomaly \,\,Score}(x_t) \in \mathbb{R}^{T \times 1}$ serves as the point-wise anomaly score for $\mathcal{X}$.

Furthermore, we conducted experiments using multi-cluster with k={1,2,3,4,5,6,7,8,9,10}. The experimental results for multi-cluster, which utilize the modified equation, are presented in Table Table~\ref{table:number_of_cluster_center_based_result}. 

\begin{table*}[h]
\vskip -0.09in
\begin{center}
\caption{Results of evaluating MADCluster performance on four real-world datasets with multi-cluster $(k=1$ to $10)$.}
\vskip +0.1in
\label{table:number_of_cluster_center_based_result}
\small
\setlength{\tabcolsep}{3.8pt}
\renewcommand{\arraystretch}{1.25}
\begin{tabular}{c|ccc|ccc|ccc|ccc}
    \hline
    \noalign{\smallskip}
    \multicolumn{1}{c|}{\textbf{Dataset}} & \multicolumn{3}{c|}{\textbf{MSL}} & \multicolumn{3}{c|}{\textbf{SMAP}} & \multicolumn{3}{c|}{\textbf{SMD}} & \multicolumn{3}{c}{\textbf{PSM}} \\
    \noalign{\smallskip}
    \hline
    \noalign{\smallskip}
    \multicolumn{1}{c|}{\textbf{\# Clusters}} & \textbf{F1} & \textbf{Aff-P} & \textbf{Aff-R} & \textbf{F1} & \textbf{Aff-P} & \textbf{Aff-R} & \textbf{F1} & \textbf{Aff-P} & \textbf{Aff-R} & \textbf{F1} & \textbf{Aff-P} & \textbf{Aff-R} \\
    \noalign{\smallskip}
    \hline
    \noalign{\smallskip}
    1  & \textbf{0.52} & \textbf{0.74} & \textbf{0.99} & \textbf{0.49} & \textbf{0.71} & 0.96 & \textbf{0.48} & \textbf{0.83} & 0.90 & \textbf{0.67} & \textbf{0.68} & \textbf{1.00} \\
    
    2  & 0.47 & 0.66 & 0.98 & 0.36 & 0.64 & \textbf{0.99} & 0.17 & 0.57 & 0.59 & 0.50 & 0.55 & \textbf{1.00} \\
    
    3  & 0.48 & 0.71 & 0.98 & 0.33 & 0.61 & \textbf{0.99} & 0.24 & 0.58 & \textbf{0.95} & 0.51 & 0.55 & \textbf{1.00}  \\
    
    4  & 0.47 & 0.69 & 0.98 & 0.35 & 0.60 & \textbf{0.99} & 0.24 & 0.61 & 0.92 & 0.50 & 0.55 & \textbf{1.00}  \\
    
    5  & 0.47 & 0.68 & \textbf{0.99} & 0.34 & 0.62 & 0.98 & 0.22 & 0.59 & 0.85 & 0.51 & 0.56 & \textbf{1.00}  \\
    
    6  & 0.49 & 0.68 & 0.98 & 0.34 & 0.58 & \textbf{0.99} & 0.23 & 0.57 & 0.92 & 0.54 & 0.57 & 0.99  \\
    
    7  & 0.49 & 0.68 & \textbf{0.99} & 0.39 & 0.65 & \textbf{0.99} & 0.28 & 0.61 & 0.88 & 0.50 & 0.55 & \textbf{1.00}  \\
    
    8  & 0.47 & 0.69 & 0.97 & 0.37 & 0.66 & 0.98 & 0.30 & 0.62 & 0.84 & 0.50 & 0.55 & \textbf{1.00} \\
    
    9  & 0.48 & 0.72 & 0.97 & 0.38 & 0.64 & 0.98 & 0.25 & 0.59 & 0.92 & 0.50 & 0.55 & \textbf{1.00}  \\
    
    10 & 0.51 & 0.71 & 0.98 & 0.37 & 0.65 & \textbf{0.99} & 0.27 & 0.63 & 0.81 & 0.62 & 0.60 & 0.69 \\
    \noalign{\smallskip}
    \hline
\end{tabular}
\end{center}
\end{table*}

Overall, across the benchmark datasets—MSL, SMAP, SMD, and PSM—the performance patterns do not consistently improve or degrade with increasing numbers of clusters. In particular, the best performance in terms of F1 and Aff-P is often observed when using a single cluster, while in the SMD and SMAP datasets, Aff-R tends to remain high across multiple clusters, although the difference from the single-cluster case is not significant. Unlike what might be expected, the number of clusters does not show a clear monotonic relationship with detection performance. That is, increasing the number of clusters does not necessarily improve anomaly detection performance. Instead, the most stable and strong performance is frequently achieved with just a single cluster.

This suggests that a single-cluster approach in MADCluster is not only sufficient to model the distribution of normal patterns across diverse time-series datasets but actually be optimal in many cases. The robustness of the MADCluster aligns with the design intent of the proposed One-directed Adaptive loss, which proves most effective when applied in this setting.

\section{Results after applying MADCluster to baseline models}
\label{appendix:Results of MADCluster apply for other models}

\subsection{Computational Efficiency}

Table~\ref{table2:computation} lists the computational costs and validation accuracy, with all models trained on the MSL dataset. When applying MADCluster, performance significantly improves without substantially impacting structural complexity or efficiency. This integration results in only a slight increase in computational demands, as measured by MACs (KMac units), with a modest increase in parameter size. By maintaining a balance between efficiency and performance, this method enhances the anomaly detection capabilities of existing models without imposing significant changes. This demonstrates the effectiveness and adaptability of MADCluster, indicating its potential to improve existing anomaly detection techniques while balancing computational demands and performance enhancement.

\begin{table}[h]
\vskip -0.15in
\caption{Computational Efficiency and metrics(F1, V\_ROC, V\_PR) Comparison on the MSL Dataset, detailing the number of parameters (`\# Params') indicating model size and Multiply-Accumulate Computations (`MACs') reflecting processing speed.}
\label{table2:computation}
\begin{center}
{\footnotesize  
\setlength{\tabcolsep}{5pt}
\renewcommand{\arraystretch}{1.1}
\begin{tabular}{c|ccccc}
    \hline
    \noalign{\smallskip}
    \textbf{Model} & \textbf{MACs} & \textbf{\#Params} & \textbf{F1} & \textbf{V\_ROC} & \textbf{V\_PR} \\ 
    \noalign{\smallskip}
    \hline
    \noalign{\smallskip}
    DeepSVDD & 31.81M & 311.55K & 0.37 & 0.63 & 0.28 \\
    \textbf{+ MADCluster} & 31.81M & 311.62K & \textbf{0.52} & \textbf{0.72} & \textbf{0.42} \\
    \noalign{\smallskip}
    \hline
    \noalign{\smallskip}
    USAD & 427.36M & 256.26M & \textbf{0.53} & 0.71 & \textbf{0.43} \\
    \textbf{+ MADCluster} & 427.36M & 256.26M & \textbf{0.53} & \textbf{0.72} & \textbf{0.43} \\
    \noalign{\smallskip}
    \hline
    \noalign{\smallskip}
    BeatGAN & 10.22G & 185.85M & 0.49 & 0.70 & 0.39 \\
    \textbf{+ MADCluster} & 10.22G & 185.85M & \textbf{0.50} & \textbf{0.71} & \textbf{0.43} \\
    \noalign{\smallskip}
    \hline
    \noalign{\smallskip}
    OmniAnomaly & 35.44M & 350.72K & 0.42 & 0.64 & 0.31 \\
    \textbf{+ MADCluster} & 35.44M & 350.83K & \textbf{0.45} & \textbf{0.67} & \textbf{0.34} \\
    \noalign{\smallskip}
    \hline
    \noalign{\smallskip}
    THOC & 69.42M & 390.78K & 0.50 & 0.71 & 0.41 \\
    \textbf{+ MADCluster} & 69.42M & 390.91K & \textbf{0.55} & \textbf{0.72} & \textbf{0.46} \\
    \noalign{\smallskip}
    \hline
    \noalign{\smallskip}
    AnomalyTransformer & 485.23M & 4.86M & 0.50 & 0.70 & \textbf{0.40} \\
    \textbf{+ MADCluster} & 485.23M & 4.86M & \textbf{0.51} & \textbf{0.71} & \textbf{0.40} \\
    \noalign{\smallskip}
    \hline
    \noalign{\smallskip}
    DCdetector & 1.189G & 912.18K & 0.28 & 0.52 & 0.19 \\
    \textbf{+ MADCluster} & 1.189G & 912.30K & \textbf{0.41} & \textbf{0.64} & \textbf{0.32} \\
    \noalign{\smallskip}
    \hline
\end{tabular}
}
\end{center}
\vskip -0.3in
\end{table}

\subsection{Impact of Clustering and Distance Mapping on Anomaly Detection Performance}

In Table~\ref{table:results_madcluster} we evaluated the performance of the anomaly detection approaches illustrated in maintext Figure~\ref{fig:Difference_between_method8}. This table presents quantitative results of our proposed method, which learns center coordinates and performs single clustering as we hypothesized. DeepSVDD represents only distance mapping, while Clustering denotes the experimental results using self-labeling without distance mapping. MADCluster, our proposed method, applies both distance mapping and clustering.

\vskip -0.09in
\begin{table*}[h]
\begin{center}
\caption{Performance comparison of anomaly detection approaches across four datasets: (1) DeepSVDD (Cluster Distance Mapping), (2) Clustering (Sequence-wise Clustering), and (3) MADCluster (Combined Cluster Distance Mapping and Sequence-wise Clustering)}
\vskip +0.1in
\label{table:results_madcluster}
\small
\setlength{\tabcolsep}{3.6pt}
\renewcommand{\arraystretch}{1.1}
\begin{tabular}{c|ccc|ccc|ccc|ccc}
    \hline
    \noalign{\smallskip}
    \multicolumn{1}{c|}{\textbf{Dataset}} & \multicolumn{3}{c|}{\textbf{MSL}} & \multicolumn{3}{c|}{\textbf{SMAP}} & \multicolumn{3}{c|}{\textbf{SMD}} & \multicolumn{3}{c}{\textbf{PSM}} \\
    \noalign{\smallskip}
    \hline
    \noalign{\smallskip}
    \multicolumn{1}{c|}{\textbf{Metric}} & \textbf{F1} & \textbf{Aff-P} & \textbf{Aff-R} & \textbf{F1} & \textbf{Aff-P} & \textbf{Aff-R} & \textbf{F1} & \textbf{Aff-P} & \textbf{Aff-R} & \textbf{F1} & \textbf{Aff-P} & \textbf{Aff-R} \\
    \noalign{\smallskip}
    \hline
    \noalign{\smallskip}
    DeepSVDD  & 0.37 & 0.63 & \textbf{}{0.99} & 0.40 & 0.70 & \textbf{0.99} & 0.19 & 0.56 & 0.61 & 0.50 & 0.55 & \textbf{1.00} \\
    
    Clustering  & 0.30 & 0.70 & 0.66 & 0.39 & 0.70 & 0.95 & 0.20 & 0.58 & 0.61 & 0.19 & 0.56 & 0.23 \\
        
    MADCluster  & \textbf{}{0.52} & \textbf{0.74} & \textbf{0.99} & \textbf{0.49} & \textbf{0.71} & 0.96 & \textbf{0.48} & \textbf{0.83} & \textbf{0.90} & \textbf{0.65} & \textbf{0.66} & 0.62 \\
    
    \noalign{\smallskip}
    \hline
    \noalign{\smallskip}
\end{tabular}
\end{center}
\vskip -0.2in
\end{table*}

Evaluation of original F1, Aff-P, and Aff-R using Cluster Distance Mapping and Sequence-wise Clustering individually did not reveal a consistently dominant method across datasets. In contrast, MADCluster, which integrates both approaches, achieved the highest performance in all metrics except Aff-R on the SMAP and PSM datasets.

\section{Extension of MADCluster to Image Data Domains}
\label{appendix:Image_Data_Domains} 

To evaluate the applicability of MADCluster beyond time-series data, experiments were conducted on image anomaly detection tasks. In the original implementation, the extracted dynamics through the Base Embedder are represented as \texttt{[batch, sequence, hidden\_dim]}. For image inputs typically structured as \texttt{[batch\_size, channels, height, width]}, the spatial dimensions were flattened to form a sequence-like representation of shape \texttt{[batch\_size, channels, height $\times$ width]}. This transformation enables the application of MADCluster sequence-wise clustering and cluster distance mapping mechanisms to spatial data.

Experiments were performed on the MVTec AD dataset~\cite{bergmann2019mvtec}, a widely used benchmark for unsupervised image anomaly detection. MADCluster was integrated with three representative models: Reverse Distillation for Anomaly Detection (RD4AD)\cite{deng2022reverse}, PyramidFlow\cite{lei2023pyramidflow}, and RealNet~\cite{zhang2024realnet}. All models were trained for 10 epochs, and performance was evaluated using the Area Under the Receiver Operating Characteristic (AUROC) at both the image and pixel levels. The results are summarized in Table~\ref{table:image-level} and Table~\ref{table:pixel-level}, demonstrating that MADCluster can be effectively extended to image domains with minimal architectural modifications.

\begin{table*}[h]
\vskip -0.1in
\caption{Image-level AUROC (\%) on the MVTec AD dataset with and without MADCluster.}
\label{table:image-level}
\begin{center}
\begin{footnotesize}
\setlength{\tabcolsep}{3.6pt}
\renewcommand{\arraystretch}{1}
\begin{tabular}{c|cc|cc|cc}
\hline
\textbf{Class Name} & \textbf{RealNet} & \textbf{+MADCluster} & \textbf{RD4AD} & \textbf{+MADCluster} & \textbf{PyramidFlow} & \textbf{+MADCluster} \\
\hline
Bottle       & 0.918 & \textbf{0.961} & 0.991 & \textbf{0.998} & 0.778 & \textbf{0.993} \\
Cable        & 0.631 & \textbf{0.669} & 0.945 & \textbf{0.946} & 0.638 & \textbf{0.695} \\
Capsule      & 0.694 & \textbf{0.698} & 0.868 & \textbf{0.870} & 0.870 & \textbf{0.916} \\
Carpet       & 0.969 & \textbf{0.977} & 0.996 & \textbf{0.997} & 0.938 & \textbf{0.964} \\
Grid         & 0.872 & \textbf{0.875} & 0.921 & \textbf{0.945} & 0.794 & \textbf{0.824} \\
Hazelnut     & 0.972 & \textbf{0.994} & \textbf{1.000} & \textbf{1.000} & 0.930 & \textbf{0.935} \\
Leather      & 0.806 & \textbf{0.830} & \textbf{1.000} & \textbf{1.000} & 0.993 & \textbf{0.999} \\
Metal Nut    & 0.670 & \textbf{0.688} & 0.995 & \textbf{0.996} & 0.735 & \textbf{0.742} \\
Pill         & 0.823 & \textbf{0.844} & 0.936 & \textbf{0.956} & 0.810 & \textbf{0.834} \\
Screw        & 0.552 & \textbf{0.572} & 0.829 & \textbf{0.848} & 0.595 & \textbf{0.752} \\
Tile         & 0.972 & \textbf{0.981} & 0.993 & \textbf{0.994} & 0.994 & \textbf{0.995} \\
Toothbrush   & 0.553 & \textbf{0.644} & 0.997 & \textbf{1.000} & 0.944 & \textbf{0.947} \\
Transistor   & 0.659 & \textbf{0.660} & 0.967 & \textbf{0.970} & 0.908 & \textbf{0.936} \\
Wood         & 0.959 & \textbf{0.966} & 0.990 & \textbf{0.993} & 0.991 & \textbf{0.996} \\
Zipper       & 0.882 & \textbf{0.901} & 0.871 & \textbf{0.889} & \textbf{0.938} & \textbf{0.938} \\
\hline
\end{tabular}
\end{footnotesize}
\end{center}
\vskip -0.15in
\end{table*}

\begin{table*}[h]
\vskip -0.1in
\caption{Pixel-level AUROC (\%) on the MVTec AD dataset with and without MADCluster.}
\label{table:pixel-level}
\begin{center}
\begin{footnotesize}
\setlength{\tabcolsep}{3.6pt}
\renewcommand{\arraystretch}{1}
\begin{tabular}{c|cc|cc|cc}
\hline
\textbf{Class Name} & \textbf{RealNet} & \textbf{+MADCluster} & \textbf{RD4AD} & \textbf{+MADCluster} & \textbf{PyramidFlow} & \textbf{+MADCluster} \\
\hline
Bottle       & 0.949 & \textbf{0.963} & 0.982 & \textbf{0.986} & 0.960 & \textbf{0.974} \\
Cable        & 0.631 & \textbf{0.897} & \textbf{0.977} & \textbf{0.977} & 0.895 & \textbf{0.912} \\
Capsule      & 0.901 & \textbf{0.927} & 0.981 & \textbf{0.982} & 0.977 & \textbf{0.980} \\
Carpet       & 0.970 & \textbf{0.984} & \textbf{0.992} & \textbf{0.992} & 0.964 & \textbf{0.978} \\
Grid         & 0.873 & \textbf{0.894} & 0.942 & \textbf{0.963} & 0.941 & \textbf{0.948} \\
Hazelnut     & 0.925 & \textbf{0.956} & \textbf{0.991} & \textbf{0.991} & 0.964 & \textbf{0.973} \\
Leather      & 0.968 & \textbf{0.971} & \textbf{0.994} & \textbf{0.994} & 0.985 & \textbf{0.987} \\
Metal Nut    & 0.754 & \textbf{0.770} & 0.969 & \textbf{0.974} & 0.938 & \textbf{0.959} \\
Pill         & 0.942 & \textbf{0.943} & 0.967 & \textbf{0.968} & 0.943 & \textbf{0.956} \\
Screw        & 0.929 & \textbf{0.946} & 0.985 & \textbf{0.986} & 0.898 & \textbf{0.903} \\
Tile         & 0.930 & \textbf{0.937} & \textbf{0.953} & \textbf{0.953} & 0.962 & \textbf{0.973} \\
Toothbrush   & 0.918 & \textbf{0.924} & 0.987 & \textbf{0.988} & 0.975 & \textbf{0.977} \\
Transistor   & 0.704 & \textbf{0.722} & \textbf{0.890} & \textbf{0.890} & 0.965 & \textbf{0.972} \\
Wood         & 0.930 & \textbf{0.932} & \textbf{0.955} & \textbf{0.955} & 0.957 & \textbf{0.960} \\
Zipper       & 0.951 & \textbf{0.962} & 0.968 & \textbf{0.970} & \textbf{0.968} & \textbf{0.968} \\
\hline
\end{tabular}
\end{footnotesize}
\end{center}
\vskip -0.15in
\end{table*}

\vspace{5mm} 
\section{Dataset}
\label{appendix:Dataset} 

We summarize the four adopted benchmark datasets for evaluation in Table~\ref{table:benchmark_datasets_appendix}. These datasets include multivariate time series scenarios with different types and anomaly ratios. MSL, SMAP, SMD and PSM are multivariate time series datasets.

\begin{table*}[h]
\vskip -0.1in
\caption{Statistics and details of the benchmark datasets used. AR (anomaly ratio) represents the abnormal proportion of the whole dataset.}
\vskip 0.05in
\label{table:benchmark_datasets_appendix}
\begin{center}
\begin{footnotesize}
\renewcommand{\arraystretch}{1.1}
\begin{tabular}{c|c|cc|cc|c}
\hline
Benchmarks & Applications & Dim & Win & \#Train & \#Test & AR (Truth) \\
\hline
MSL & Space & 55 & 100 & 58,317 & 73,729 & 0.105 \\
SMAP & Space & 25 & 100 & 135,183 & 427,617 & 0.128 \\
SMD & Server & 38 & 100 & 708,405 & 708,420 & 0.042 \\
PSM & Server & 25 & 100 & 132,481 & 87,841 & 0.278 \\
\hline
\end{tabular}
\end{footnotesize}
\end{center}
\vskip -0.15in
\end{table*}

\end{document}